\documentclass[10pt,journal,cspaper,compsoc]{IEEEtran}
\usepackage{times}
\usepackage{graphicx}
\usepackage{paralist}
\usepackage{epstopdf}
\usepackage{verbatim}
\usepackage{epsfig}
\usepackage{amsmath}
\usepackage{amssymb}
\usepackage{amsfonts}
\usepackage{color}
\usepackage[square,numbers]{natbib}
\usepackage{multirow}
\usepackage{rotating}
\usepackage{stfloats}
\usepackage{ifthen}
\usepackage{times}
\usepackage{dsfont}
\usepackage{booktabs}
\usepackage{subfigure}
\usepackage{soul}
\usepackage{makecell}
\usepackage{pifont}

\usepackage[breaklinks,colorlinks]{hyperref}

\usepackage[ruled,vlined]{algorithm2e}
\usepackage{xcolor}
\definecolor{commentcolor}{RGB}{110,154,155}   
\newcommand{\PyComment}[1]{\ttfamily\textcolor{commentcolor}{\# #1}}  
\newcommand{\PyCode}[1]{\ttfamily\textcolor{black}{#1}} 

\usepackage{url}

\newcommand{\bz}{\mathbf{z}}
\newcommand{\bq}{\mathbf{q}}
\newcommand{\bk}{\mathbf{k}}
\newcommand{\bv}{\mathbf{v}}
\newcommand{\bU}{\mathbf{U}}
\newcommand{\real}{\mathbb{R}}

\newcommand{\CUT}[1]{}

\newcommand{\jimmy}{}
\newcommand{\jimmyy}{}

\usepackage[capitalize]{cleveref}
\crefname{section}{Sec.}{Secs.}
\Crefname{section}{Section}{Sections}
\Crefname{table}{Table}{Tables}
\crefname{table}{Tab.}{Tabs.}

\newfont{\bboard}{msbm10 scaled\magstephalf}

\def\real{\mbox{\bboard R}}

\def\independenT#1#2{\mathrel{\setbox0\hbox{$#1#2$}%
\copy0\kern-\wd0\mkern4mu\box0}}

\usepackage{silence}
\begin{document}

\title{DropMAE: \jimmyy{Learning Representations via Masked Autoencoders with Spatial-Attention Dropout for Temporal Matching Tasks}}

\author{Qiangqiang~Wu, Tianyu~Yang, Ziquan~Liu, Wei~Lin, Baoyuan~Wu and Antoni~B.~Chan 
\IEEEcompsocitemizethanks{\IEEEcompsocthanksitem Qiangqiang~Wu, Wei~Lin and Antoni~B. Chan (corresponding author)  are with the Department of Computer Science, City University of Hong Kong.
E-mail: qiangqwu2-c@my.cityu.edu.hk, wlin38-c@my.cityu.edu.hk, abchan@cityu.edu.hk.
\IEEEcompsocthanksitem Tianyu Yang is with Alibaba DAMO Academy, HangZhou, China. (e-mail: tianyu-yang@outlook.com)
\IEEEcompsocthanksitem Ziquan Liu is with School of Electronic Engineering and Computer Science, Queen Mary University of London. (e-mail: ziquanliu.cs@gmail.com)
\IEEEcompsocthanksitem Baoyuan Wu is with School of Data Science, The Chinese University of Hong Kong, Shenzhen, Guangdong, 518172, P.R. China. (e-mail: wubaoyuan@cuhk.edu.cn)
}
\thanks{}}

\markboth{Journal of \LaTeX\ Class Files,~Vol.~X, No.~X, XXX~XXXX}%
{Shell \MakeLowercase{\textit{et al.}}: Bare Demo of IEEEtran.cls for Computer Society Journals}

\IEEEcompsoctitleabstractindextext{%
\begin{abstract}
This paper studies masked autoencoder (MAE) video pre-training for various temporal matching-based downstream tasks, i.e., object-level tracking tasks including video object tracking (VOT) and video object segmentation (VOS), self-supervised visual correspondence learning, dense tracking tasks including optical flow estimation and long-term point tracking, and 3D point cloud tracking. Specifically, our work explores to provide a general representation to boost the temporal matching ability in various downstream tracking tasks. To achieve this, we firstly find that a simple extension of MAE, which randomly masks out frame patches in videos and reconstruct the frame pixels, heavily relies on spatial cues while ignoring temporal relations for frame reconstruction, thus leading to sub-optimal temporal matching representations. To alleviate this, we propose DropMAE, which adaptively performs spatial-attention dropout in the frame reconstruction to facilitate temporal correspondence learning in videos. We obtain several important findings with DropMAE: 1) DropMAE is a strong and efficient temporal {matching} learner, which achieves better fine-tuning results on matching-based tasks than the ImageNet-based MAE with {$2\times$ faster pre-training speed}. 2) DropMAE is effective for different tracking tasks, i.e., object-level matching tasks including VOT and VOS, dense tracking tasks including optical flow estimation and tracking any point (TAP), and even 3D tracking in the different modality of point cloud data. 3) DropMAE can significantly speed up (e.g., 16.6$\times$ faster on K400) the existing self-supervised visual correspondence learning by using the DropMAE pre-trained weights. 4) Motion diversity in pre-training videos is more important than scene diversity for improving the downstream tracking performance. Since none exists, we build ViT-based trackers for different downstream tracking tasks, and our pre-trained DropMAE model can be directly loaded in these ViT-based trackers for fine-tuning without further modifications. 
Experiments on 6 downstream tracking tasks demonstrate the effectiveness of DropMAE as a general pre-trained representation for diverse tracking tasks. The code and pre-trained models are available at \url{https://github.com/jimmy-dq/DropMAE.git}.
\end{abstract}

\begin{IEEEkeywords}Generative Pre-training, Video Object Tracking, Video Object Segmentation, Self-supervised Correspondence Learning, Optical Flow Estimation, Long-term Point Tracking, Deep Learning
\end{IEEEkeywords}
}

\maketitle

\IEEEdisplaynotcompsoctitleabstractindextext
\IEEEpeerreviewmaketitle

\section{Introduction}
\CUT{
Recently, transformers have achieved enormous  success in many research areas, such as natural language processing (NLP) \cite{gpt_1,BERT}, computer vision \cite{ostrack} and audio generation \cite{audio_generation,Neural_speech_synthesis}. In NLP, masked autoencoding is commonly used to train large-scale generalizable NLP  transformers containing billions of parameters. Inspired by the great success of self-supervised learning in NLP, recent advances \cite{mae,SimMIM} in computer vision suggest that training large-scale vision transformers may undertake a similar trajectory with NLP. The seminal work MAE \cite{mae} reconstructs the input image from a small portion of patches. The learned representation in this masked autoencoder has been demonstrated to be effective in many computer vision tasks, such as image classification, object detection and semantic segmentation.}

Transformers have achieved remarkable success across various research areas, including natural language processing (NLP) \cite{gpt_1,BERT}, computer vision \cite{ostrack}, and audio generation \cite{audio_generation,Neural_speech_synthesis}. In NLP, masked autoencoding has become a standard approach for training large-scale transformers with billions of parameters, enabling robust and generalizable representations. Building on the success of self-supervised learning in NLP, recent studies in computer vision \cite{mae,SimMIM} have explored similar strategies for vision transformers. Among these, the pioneering work MAE \cite{mae} introduces a masked autoencoding framework that reconstructs input images from a subset of patches. The learned representations from MAE have demonstrated effectiveness across a variety of computer vision tasks, including image classification, object detection, and semantic segmentation.

\CUT{
In video object tracking (VOT), recently two works,  SimTrack \cite{simtrack} and OSTrack \cite{ostrack}, explore  using an MAE pre-trained ViT model as the tracking backbone. Notably, these two trackers achieve state-of-the-art performance on existing tracking benchmarks without using {complicated tracking pipelines.} 
The key to their success is the {robust} pre-training weights learned by MAE on ImageNet \cite{ILSVRC}. In addition, \cite{simtrack,ostrack} also demonstrate that, for VOT, MAE {\emph{unsupervised}} pre-training on ImageNet is more effective than {\emph{supervised}} pre-training using class labels -- this is mainly because MAE pre-training learns more fine-grained local structures that are useful for accurate target localization required for VOT, {whereas supervised training learns high-level class-related features that are invariant over appearance changes}. Despite the promising performance achieved by \cite{simtrack,ostrack}, the MAE pre-training on ImageNet could still be sub-optimal for the tracking task due to the natural gap between images and videos, i.e., no prior temporal correspondence information can be learned in static images. However, previous tracking methods \cite{pul,SiamFC,siamrpn} have shown that temporal correspondence learning is the key in developing a robust and discriminative tracker.  {Thus there is an opportunity to further develop the MAE framework specifically for matching-based video tasks, such as VOT and VOS.}
}

Recent advancements in video object tracking (VOT) have demonstrated the effectiveness of leveraging MAE pre-trained ViT models as backbones, as evidenced by SimTrack \cite{simtrack} and OSTrack \cite{ostrack}. Notably, these two trackers achieve state-of-the-art performance on standard benchmarks without using complicated tracking pipelines. Their success can be attributed to the robust pre-trained weights obtained from MAE on ImageNet \cite{ILSVRC}. Moreover, \cite{simtrack,ostrack} demonstrate that, for VOT, MAE's \emph{unsupervised} pre-training on ImageNet surpasses \emph{supervised} pre-training with class labels -- this is mainly because MAE pre-training effectively captures fine-grained local structures essential for precise target localization in VOT, whereas supervised pre-training focuses on high-level, class-specific features that are invariant to appearance changes and less suited for tracking tasks. Despite the strong performance of \cite{simtrack,ostrack}, the MAE pre-training on ImageNet remains sub-optimal for tracking tasks due to the natural gap between images and videos, i.e., no prior temporal correspondence information can be learned in static images. Moreover, previous methods \cite{pul,SiamFC,siamrpn} have demonstrated that learning temporal correspondences is essential for building robust and discriminative trackers. Thus there is an opportunity to further develop the MAE framework specifically for matching-based downstream tracking tasks, such as VOT, VOS and dense tracking tasks (e.g., optical flow estimation and long-term point tracking).

 \begin{figure}
\begin{center}
   \includegraphics[width=1.0\linewidth]{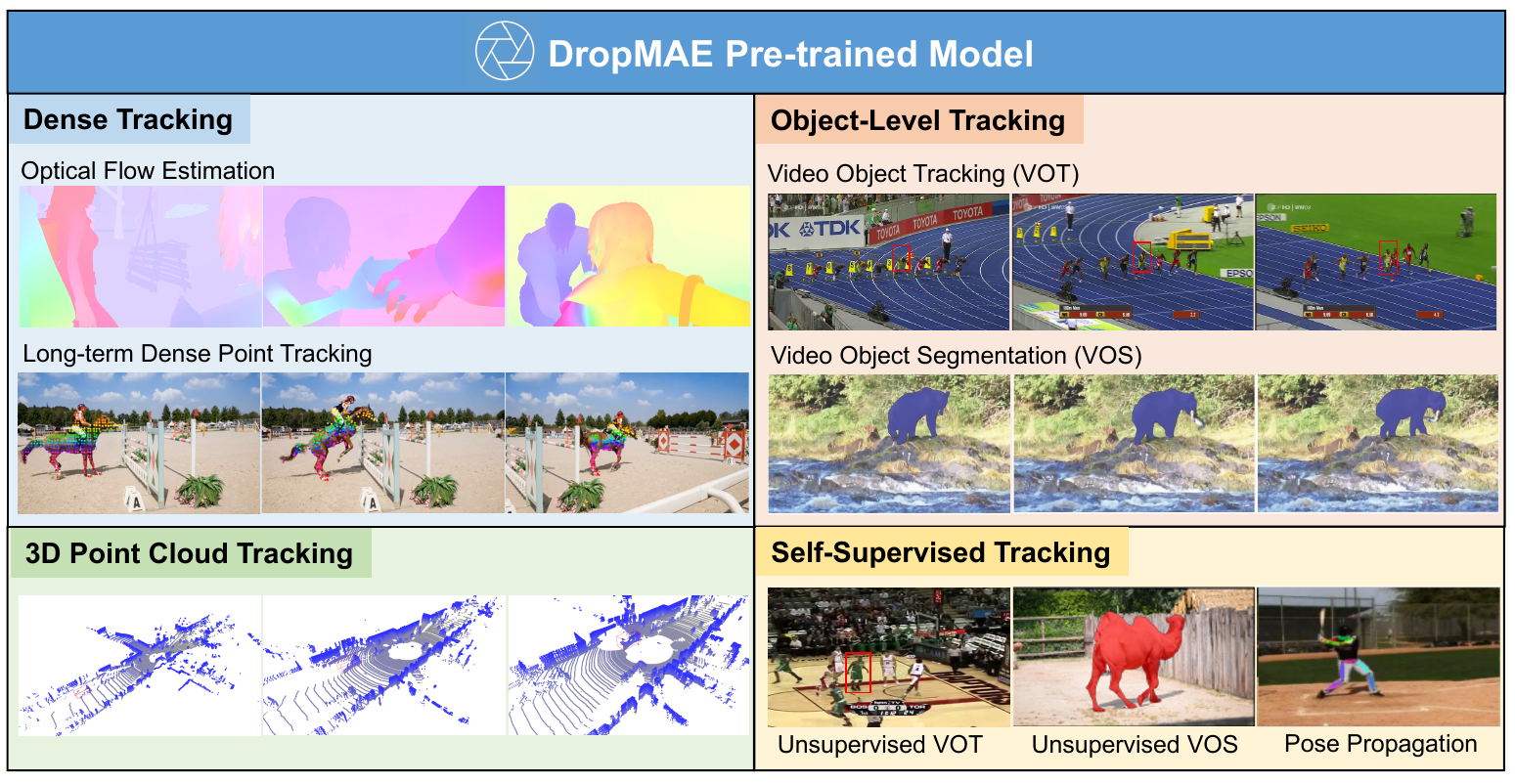} 
\end{center}
\vspace{-0.55cm}
 \caption{A general DropMAE pre-trained model for various downstream tracking tasks including object-level tracking (i.e., VOT and VOS), 3D point cloud tracking, dense tracking (i.e., optical flow estimation and long-term point tracking) and self-supervised correspondence learning for unsupervised tracking.
 }
\label{unified_model}
\end{figure}

\begin{figure}
\begin{center}
   \includegraphics[width=0.85\linewidth]{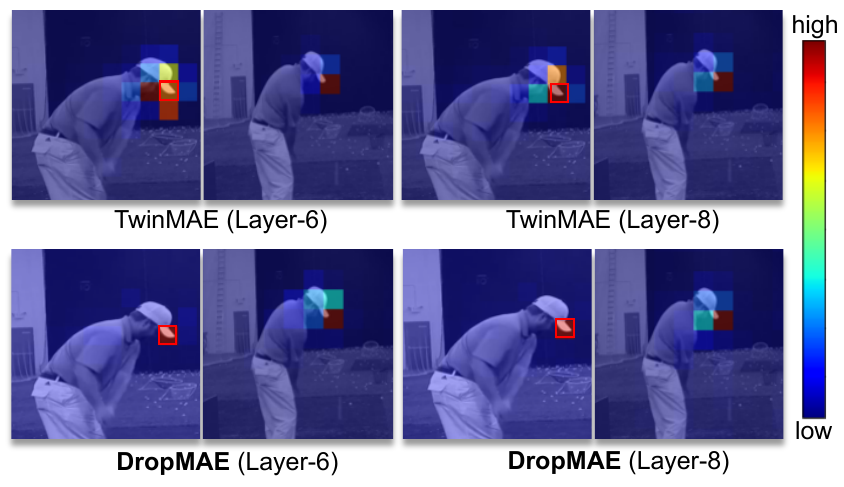} 
\end{center}
\vspace{-0.65cm}
 \caption{Visualization of the attention maps of the TwinMAE baseline { and our DropMAE} in the reconstruction of a random masked patch, {which is denoted as a red bounding box in the left input frame}. TwinMAE 
 {leverages the spatial cues (within the same frame) more than temporal cues (between frames) for reconstruction.}
 Our proposed DropMAE  improves the baseline by effectively alleviating co-adaptation between spatial cues in the reconstruction, {focusing more on temporal cues}, thus achieving better learning of temporal correspondences for tracking tasks.
 }
\label{att_temp_mae}
\end{figure}

\CUT{
One simple way to extend MAE to videos is to randomly mask out \CUT{{random 2D}}{frame}
patches 
in a video clip (i.e., video frame pairs) and then reconstruct the video clip.
We denote this simple baseline as \emph{twin MAE} {(TwinMAE)}.
Given a masked patch query, as illustrated in {Figs.~\ref{att_temp_mae} \& \ref{temp_spatial_attention_comp}}, we find that  TwinMAE heavily relies on the spatially neighbouring patches \emph{within the same frame} to reconstruct the masked patch, which implies a heavy co-adaptation of spatial cues (within-frame tokens) for reconstruction and may cause learning of sub-optimal temporal representations for matching-based downstream tasks like video object tracking and segmentation.}

The naive extension of MAE to video representation learning involves randomly masking patches across frames in a video clip (e.g., frame pairs) and reconstructing the clip. We denote this simple baseline as \emph{twin MAE} {(TwinMAE)}. As illustrated in {Figs.~\ref{att_temp_mae} \& \ref{temp_spatial_attention_comp}}, given a masked patch query, TwinMAE primarily relies on spatially adjacent patches within the same frame for reconstruction, which implies a heavy co-adaptation of spatial cues (within-frame tokens) for reconstruction and may cause learning of sub-optimal temporal representations for matching-based downstream tasks like video object tracking and segmentation. 


\CUT{
To address this issue with the TwinMAE baseline, we propose DropMAE specifically designed for pre-training a masked autoencoder for matching-based video downstream tasks (e.g., VOT and VOS). Our DropMAE adaptively performs spatial-attention dropout to break up co-adaptation between spatial cues (within-frame tokens) during the frame reconstruction, thus encouraging more temporal interactions and facilitating temporal correspondence learning in the pre-training stage. 
Interestingly, we obtain several important findings with DropMAE: 1) DropMAE is an effective and efficient temporal correspondence learner, which achieves better fine-tuning results on matching-based tasks than the ImageNet-based MAE with {$2\times$ faster pre-training speed}. 2) Motion diversity in pre-training videos is more important than scene diversity for improving the performance on VOT and VOS.}

To solve the aforementioned issue with the TwinMAE baseline, we propose DropMAE, a pre-training method designed for masked autoencoders in temporal matching-based video downstream tasks (e.g., VOT, VOS, self-supervised correspondence learning and long-term dense tracking). Our DropMAE adaptively performs spatial-attention dropout to break up co-adaptation between spatial cues (within-frame tokens) during the frame reconstruction, which encourages stronger temporal interactions and facilitates the learning of temporal correspondences during pre-training. Interestingly, we observe several key findings with DropMAE: 1) DropMAE is a strong and efficient temporal {matching} learner, which achieves better fine-tuning results on matching-based tasks than the ImageNet-based MAE with {$2\times$ faster pre-training speed}. 2) Motion diversity in pre-training videos is more important than scene diversity for improving the downstream tracking performance. 3) DropMAE is effective for different tracking tasks, i.e., object-level tracking including VOT and VOS, pixel-level based dense tracking tasks including optical flow estimation and long-term point tracking, self-supervised correspondence learning in videos and even tracking in the different modality of 3D point cloud data. 4) Leveraging DropMAE as pre-trained weights results in a $16.6\times$ speedup for self-supervised visual correspondence learning on K400.


To evaluate the effectiveness of our DropMAE, we conduct experiments on 6 downstream tracking tasks, including \emph{VOT}, \emph{VOS}, \emph{self-supervised visual correspondence learning}, dense tracking including \emph{optical flow estimation} and \emph{long-term point tracking}, and \emph{3D point cloud tracking}. Since some downstream tasks may lack ViT-based tracking baselines, we build these ViT baselines for further study. For VOT and VOS, we find that our trackers with DropMAE pre-training obtain 75.9\% AO on GOT-10k, 52.7\% AUC on LaSOT$_{ext}$, 56.9\% AUC on TNL2K and {92.1}\%/83.0\% $\mathcal{J}\&\mathcal{F}$ scores on DAVIS-16/17, w/o using complicated online updating or memory mechanisms. For long-term point tracking, our test-time optimization based DropDINO tracker achieves a new state-of-the-art AJ score of 65.6\% on DAVIS-480 \cite{davis17}, outperforming DINO-Tracker \cite{tumanyan2025dino} w/ fewer learnable parameters. For 3D point cloud tracking, we show that our DropMAE performs favourably against 3D pre-training approaches while significantly outperforming 2D MAE, showing its potential in 3D tracking. Overall, the competitive tracking performance achieved by various downstream DropMAE trackers show the superiority of our pre-training approach.


\CUT{
{We conduct downstream task evaluation on 9 competitive VOT and VOS benchmarks, achieving state-of-the-art performance on these benchmarks. In particular, our trackers with DropMAE pre-training obtain 75.9\% AO on GOT-10k, 52.7\% AUC on LaSOT$_{ext}$, 56.9\% AUC on TNL2K and {92.1}\%/83.0\% $\mathcal{J}\&\mathcal{F}$ scores on DAVIS-16/17, w/o using complicated online updating or memory mechanisms.}}

In summary, the main contributions of our work are:
\begin{compactitem}
  \item 
To the best of our knowledge, we are the first to investigate masked autoencoder video pre-training for various temporal matching-based downstream tasks, i.e., object-level tracking tasks like video object tracking and video object segmentation,   pixel-level dense tracking tasks like optical flow estimation and long-term point tracking, self-supervised visual correspondence learning and 3D point cloud tracking. 
 \item
We explore various video data sources for pre-training and build a TwinMAE baseline to study its effectiveness on various  temporal matching tasks. 
\item
Since some downstream tasks lack ViT-based tracking baselines, we build ViT baselines for the tasks including video object segmentation, optical flow estimation, long-term point tracking, and self-supervised visual correspondence learning. The built ViT baselines can directly load our DropMAE pre-trained model for fine-tuning without further modifications.
  \item
 We propose DropMAE, which adaptively performs spatial-attention dropout in the frame reconstruction to facilitate effective temporal correspondence learning in videos. The pre-trained DropMAE model can be directly loaded in our ViT-based downstream trackers for fine-tuning without further modifications.
 \item
 Experiments on 6 downstream tracking tasks across 13 benchmarks demonstrate the effectiveness of DropMAE as a general pre-trained representation for diverse tracking tasks. 
 \end{compactitem}

 \jimmy{A preliminary version of our work appears in our conference paper \cite{dropmae}, which considered only two downstream tasks, VOT and VOS. In this paper, we extend over the conference version by investigating 4 more downstream tasks as follows: 1) we extend DropMAE to dense tracking by proposing DropRAFT and DropDINO trackers for optical flow estimation and long-term point tracking, respectively; 2) We investigate DropMAE for self-supervised visual correspondence learning and integrate it into existing self-supervised approaches, achieving a substantial speedup in training (e.g., 16.6$\times$ faster on K400) while achieving competitive unsupervised tracking performance; 3) We apply DropMAE for 3D point cloud tracking, demonstrating its temporal matching ability generalizes to a different modality; 4) We include new experiments on dense tracking, self-supervised correspondence learning and 3D point cloud tracking.}
 
 
 \jimmy{The remainder of this paper is organized as follows. The relevant works are reviewed in \S\ref{sec:related}. We then introduce the proposed methods in \S\ref{sec:method}.  Following that, \S\ref{sec:exp} shows the comparison between our DropMAE and existing pre-training methods. In \S\ref{sec:sota_comparison}, the experimental results on 6 downstream tracking tasks are presented and discussed, while \S\ref{sec:abl} presents the ablation studies. Finally, we conclude the paper in \S\ref{sec:conclusion}.}

\section{Related Works}\label{sec:related}
In this section, we 
review the relevant works of video object tracking and segmentation, self-supervised learning, 3D point cloud tracking, optical flow estimation and long-term point tracking.
\subsection{Video Object Tracking and Segmentation.}
Given an annotated bounding box in the first frame of a test video, video object tracking (VOT) aims to accurately predict the target's bounding boxes in the following frames. Similarly, for visual object segmentation (VOS), {given an annotated binary mask in the first frame,} VOS aims to predict dense target masks in the remaining frames.
In the early development of VOT, correlation filter-based approaches 
\cite{meta_graph,new_frame_cf,BACF,KCF,robust_cf_journal} were dominant trackers due to their favorable ability in {modeling target appearance variation}. 
With the development of deep learning, deep Siamese networks \cite{SINT++} were introduced to VOT. The representative work SiamFC \cite{SiamFC} takes template and search images as input for target localization. Based on SiamFC, many improvements have been made, e.g., scale regression \cite{siamrpn,SiamRPN_plus}, online template updating \cite{memtrack,updatenet}, multi-level feature fusion \cite{CRPN}, and backbone design \cite{ostrack,simtrack,SiamRPN_plus,siamdw}. For VOS, matching-based approaches, e.g., STM \cite{STM}, AOT \cite{AOT} and STCN \cite{STCN}, achieve promising results on existing VOS benchmarks. 
Recent improvements \cite{XMEM,SWEM} on online memory design further improve the previous SOTA results in VOS. Recent studies like SimTrack \cite{simtrack} and OSTrack \cite{ostrack} show that the ViT backbone \cite{ViT} with MAE pre-training {on ImageNet} is effective for object tracking. 

Despite the great success of ViT with MAE pre-training on tracking, this static ImageNet-based pre-training still lacks temporal correspondence learning. Moreover, the developments of VOT and VOS show that learning strong temporal matching ability is essential for video tracking tasks. To the best of our best knowledge, we are the first to investigate  masked autoencoder self-supervised video pre-training for \emph{tracking} tasks. Our 
DropMAE can provide robust pre-trained weights for tracker initialization, which has been demonstrated in existing trackers, e.g., HIPTrack \cite{hiptrack}, DiffusionTrack \cite{diffusiontrack}, TGTrack \cite{shi2025historical} and BofN \cite{BofN}.


\subsection{Self-Supervised Learning} Self-supervised learning has received significant interest in the past few decades. There are many manually-designed pretext tasks for pre-training, such as image colorization \cite{coloring}, jigsaw puzzle solving \cite{jigsaw}, future frame prediction \cite{future_prediction_1,future_prediction_2} and rotation prediction \cite{rotation_pred}. Contrastive learning approaches \cite{pic,simclr,discri_ufl,moco,membank,pixel_cl} are the mainstream self-supervised methods in recent years. Concurrent works \cite{feichtenhofer2021large,bai2020can,dave2022tclr,han2020self,qian2021spatiotemporal,DUL} have demonstrated that incorporating temporally-invariant constraints into contrastive learning-based approaches enhances video action recognition performance. However, these contrastive  learning-based methods are sensitive to {the type and strength of applied data augmentation}, 
which makes them hard to train. Inspired by masked language modeling \cite{BERT,gpt_1}, masked image modeling (MIM) approaches are proposed for learning unsupervised image  \cite{mae,SimMIM}. \jimmyy{VideoMAE \cite{videomae} and ST-MAE \cite{videomae_kaiming} perform cube masking to learn video representations, which have been shown to be effective for many high-level downstream tasks including image classification and video action recognition.} SiamMAE \cite{gupta2023siamese} extends MAE \cite{mae} by masking the future frame to perform the single task of self-supervised temporal correspondence learning. \jimmyy{However, these video-based generative approaches still lack adaptive masking strategies for more effective learning of temporal matching. Moreover,} there are no specifically designed general pre-training approaches for current 
temporal matching-based tracking task.
 In this work, we propose DropMAE to explore this direction. \jimmyy{Our DropMAE leverages the adaptive spatial-attention dropout to better facilitate the
temporal learning.} As a general pre-training model, DropMAE demonstrates strong self-supervised learning capabilities for temporal matching.

\subsection{3D Point Cloud Tracking}
Inspired by 2D VOT, 3D single object tracking (SOT) aims to estimate 3D target bounding boxes in the point cloud data given the initial 3D target bounding box in the first frame. P2B \cite{qi2020p2b} proposes to employ a 3D region proposal network to generate 3D proposals, which achieves better performance on \cite{geiger2013vision,caesar2020nuscenes,waymo} than the previous shape completion-based SC3D \cite{sc3d} while running at the real-time speed. To leverage the box prior, BAT \cite{bat} extends P2B with a box-aware feature design. V2B \cite{v2b} further proposes to perform 3D SOT in the Bird’s Eye View (BEV). However, these approaches may suffer from performance degradation in challenging scenes due to the limited matching ability. To alleviate this, 3D transformer-based approaches are proposed for target-aware feature learning. The improvements include advanced transformer architectures \cite{lttr,pttr,cmt,stnet,feng2023multi,wu2022multi}, online memory modeling \cite{cxtrack,mbptrack,tat}, motion prediction \cite{m2track,dmt} and 2D-to-3D distillation \cite{wu2023boosting}. Despite these successes, pre-training research in 3D SOT remains limited. In this work, we demonstrate that the DropMAE pre-trained mode learned from 2D videos can improve 3D tracking performance when the 3D point cloud training data is limited.


\subsection{Optical Flow Estimation}
Optical flow estimation is a fundamental task that estimates per-pixel 2D motion between video frames. Traditional approaches aim to maximize visual similarity between corresponding pixels with strong regularization \cite{chen2016full,zach2007duality,horn1981determining}. Deep learning-based approaches solve the task in an end-to-end trainable manner. Specifically,  FlowNets \cite{flownet,flownet2.0} formulates the task as a dense regression problem. DCNet \cite{xu2017accurate} and PWC-Net \cite{sun2018pwc} introduce a 4D cost volume to explicitly model pixel correspondences in videos. RAFT \cite{raft} improves previous approaches by applying iterative recurrent refinements on a multi-scale 4D cost volume, which inspires numerous follow-up works \cite{flowformer,jahedi2024ms,luo2022learning,luo2023gaflow,sui2022craft,sun2022skflow,zhao2022global}. SEA-RAFT \cite{searaft} introduces  a more efficient and accurate RAFT for optical flow. 
FlowFormer++ \cite{shi2023flowformer++} introduces masked cost-volume autoencoding, leveraging MAE \cite{mae} pre-training specifically for optical flow estimation by masking the cost volume. In contrast, our DropMAE employs adaptive video frame masking, offering broader applicability across various downstream tracking tasks. Notably, DropMAE achieves a lower average endpoint error (AEPE) on the Sintel \cite{sintel} benchmark's final pass with synthetic data fine-tuning, demonstrating superior generalization performance compared to FlowFormer++.



\subsection{Long-term Point Tracking}

Previous optical flow methods focus on pixel displacement but struggle with long-term tracking consistency. Recent works address this by leveraging advanced models and datasets. PIPs \cite{harley2022particle} introduces an MLP-Mixer for iterative track updates, while TAP-Net \cite{doersch2022tap} provides a point track dataset and a neural network for location regression. TAPIR \cite{doersch2023tapir} refines point trajectories using the MLP-Mixer from PIPs. MFT \cite{neoral2024mft} selects reliable flow chains for long-term tracking by analyzing flow uncertainty and occlusion. PointOdyssey \cite{zheng2023pointodyssey} enhances temporal tracking with PIPs++, using temporal convolution. Transformer-based methods \cite{karaev2025cotracker,karaev2024cotracker3,li2025taptr,cho2025local,shrivastava2024self,aydemir2024can} further improve tracking reliability. However, these methods rely on offline training with synthetic datasets \cite{doersch2022tap,zheng2023pointodyssey}, lacking online adaptation for specific test videos.


To adapt to online testing videos, several test-time optimization approaches are proposed. 
OmniMotion \cite{wang2023tracking} refines point tracks by lifting 2D pixels to 3D. DecoMotion \cite{li2025decomposition} enhances this by decomposing videos into static and dynamic components. To speed up OmniMotion, \cite{song2025track} introduces CaDeX++ to factorize spatial-temporal features. Recently, DINO-Tracker \cite{tumanyan2025dino} uses test-time training with the pre-trained DINO-ViT model \cite{oquab2023dinov2}, achieving SOTA point tracking performance. 
In this work, we show that our DropMAE is effective for test-time optimization in online videos with limited training data due to its robust pre-trained weights. With fewer fine-tuned parameters than existing methods, our DropMAE tracker sets new SOTA tracking performance.



\section{Method}\label{sec:method}
\begin{figure}
\begin{center}
   \includegraphics[width=1.0\linewidth]{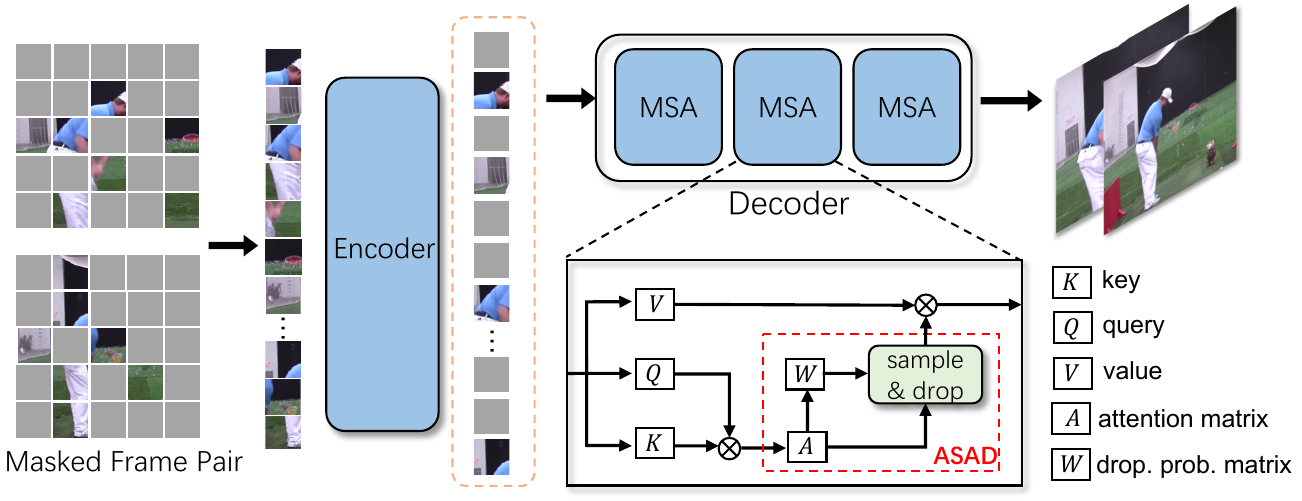} 
\end{center}
\vspace{-0.55cm}
 \caption{{An illustration of our DropMAE. The proposed adaptive spatial-attention dropout (ASAD) facilitates temporal correspondence learning for temporal matching tasks. TwinMAE follows the same pipeline except that the ASAD module is not used.}}
\label{pipeline}
\end{figure}

We propose a self-supervised video pre-training method {to learn} robust representations for temporal matching-based downstream tasks, including video object tracking (VOT), video object segmentation (VOS), 3D point cloud tracking, optical flow estimation, long-term point tracking and self-supervised visual correspondence learning. 
We firstly introduce a {simple extension of MAE to temporal matching representation learning from video,}\CUT{simple baseline using MAE to learn spatiotemporal representations in frame pairs of videos,}  denoted as the TwinMAE baseline. We then illustrate the limitations of TwinMAE and propose a spatial-attention dropout strategy to facilitate  temporal correspondence learning, denoted as DropMAE. {The overall pipeline of both DropMAE and TwinMAE is shown in Fig. \ref{pipeline}.} Finally, we introduce the various ViT-based baselines used for fine-tuning various downstream tracking tasks. 

\subsection{TwinMAE: Temporal Masked Autoencoder Baseline}
\label{temporalmae}
The masked autoencoder (MAE) model \cite{mae} consists of an encoder and a decoder. The basic idea is to randomly mask out a large portion (e.g., 75\%) of patches in an image and then reconstruct the image pixels. Specifically, the encoder only takes visible patches as input for feature learning, and then the decoder is input with both visible and masked patches to produce the image reconstruction.
In order to adapt to downstream video matching-based tasks, one naive extension is to directly apply MAE on {concatenated} video {frames}, {hoping to learn} {temporal matching}
representations from video frame pairs, which we denote as TwinMAE.

It should be noted that existing works \cite{videomae_kaiming,videomae} that extend MAE to video {representation} learning are mainly designed for the downstream task of video action recognition, where a long video clip (e.g., 16 frames) is used {for} reconstruction-based pre-training. To keep consistent with our downstream tracking tasks, we follow the general training settings used in object tracking \cite{ostrack,SiamFC} and dense tracking \cite{raft,tumanyan2025dino}        , where  two frames are sampled from {one} video as input to TwinMAE for pre-training. This adaptation significantly reduces the computational and memory cost  compared to existing video pre-training approaches \cite{videomae_kaiming,videomae}, due to the quadratic complexity of ViTs. 

\noindent\textbf{Patch embedding.} Firstly, we randomly sample 2 frames within a video with a predefined maximal frame gap. For each frame, we follow the vanilla ViT to divide it into non-overlapping patches. The patches extracted from the two frames are then concatenated together to form the overall patch sequence. We then randomly mask out patches in the patch sequence until a predefined mask ratio is reached. Note that we use the same mask ratio (i.e., 75\%) with the original MAE, since the information redundancy of two frames should be similar to a single image. The visible patches are embedded by linear projection \cite{ViT}, and the masked patches are embedded using a shared learnable mask token. All the embedded patches are added with positional embeddings \cite{ViT}.

\noindent\textbf{{Frame identity}\CUT{Temporal positional} embedding.} To distinguish between the masked tokens in the same spatial location of the two frames, we use two learnable \CUT{temporal positional }{frame identity} embeddings to indicate the two input frames. 
{The corresponding frame identity embedding is added to each embedded patch.}



\noindent\textbf{Autoencoder and Training.} Following the autoencoding pipeline in the original MAE \cite{mae}, the encoder only takes visible embedded patches as input, and the decoder is input with all the embedded patches for masked patch reconstruction. We use the same normalized pixel loss from MAE for training the whole network.


\subsection{Limitation of TwinMAE Baseline}
\label{text:limitTempMAE}
\begin{figure}
\begin{center}
   \includegraphics[width=0.95 \linewidth]{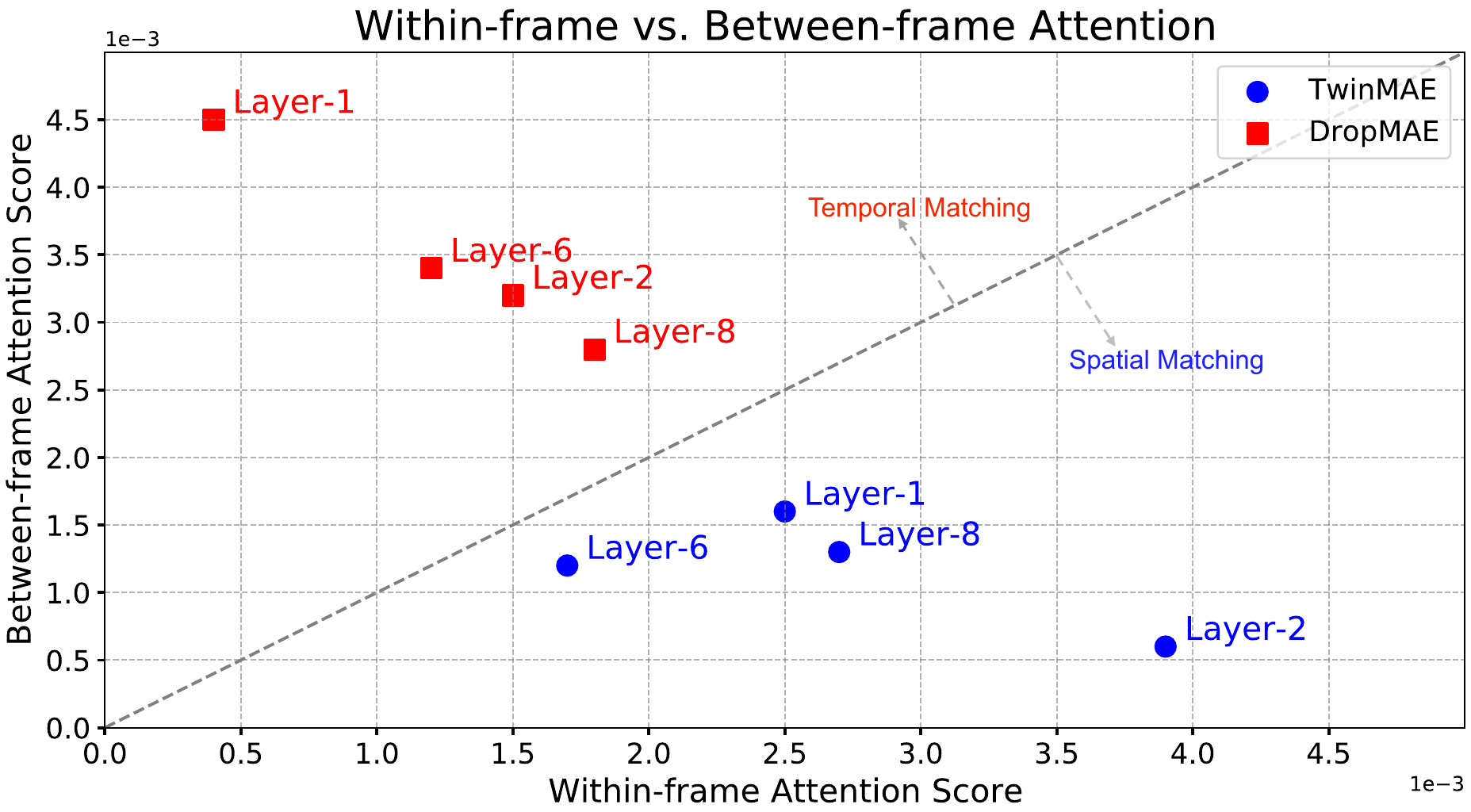} 
\end{center}
\vspace{-0.55cm}
 \caption{{The average within-frame and between-frame attention scores obtained by TwinMAE and DropMAE in different decoder layers. The attention score is calculated on 20 randomly sampled K400 validation videos, and is averaged on all heads and locations.}
 }
\label{temp_spatial_attention_comp}
\end{figure}

The visualization of the reconstruction for our TwinMAE baseline is shown in Fig.~\ref{att_temp_mae}. {We also quantitatively compare the average within-frame and between-frame attentions during the reconstruction in Fig.~\ref{temp_spatial_attention_comp}. 
 Interestingly, we find the TwinMAE reconstruction heavily relies on \emph{within-frame} patches or spatial cues, which may lead to sub-optimal temporal representations for matching-based video tasks. }
%
%
%
%
{When only using within-frame spatial cues, the decoder will perform the reconstruction using only context information in the neighboring patches, and thus the learned encoder representations will embed context information. In contrast, when using between-frame cues, the decoder will learn to perform matching of  patches \emph{between} frames so as to recover the corresponding target patch in the other frame. 
Thus, decoding with between-frame cues will make the encoder learn representations that support temporal matching between frames.}
Previous works in VOT \cite{pul,SiamFC} also suggest that temporal correspondence learning plays a key role in developing a robust and discriminative tracker. Since TwinMAE relies more on context information,  it is still suboptimal for downstream tracking tasks. 

\subsection{DropMAE via Adaptive Spatial-Attention Dropout}

To address issue of TwinMAE discussed in \S\ref{text:limitTempMAE}, we propose an 
\emph{Adaptive Spatial-Attention Dropout} (ASAD) to facilitate the temporal correspondence learning in the model, which we denote as DropMAE. 
Given a query token, our basic idea is to adaptively drop a portion of its within-frame cues in order to facilitate the model to learn more reliable temporal correspondence, i.e., between-frame cues. 
That is, we restrict the interactions between the query token and tokens in the same frame, and encourage more interactions with tokens in the other frame, {through manipulation of the computed spatial-attention in the transformer.} Therefore, to minimize the reconstruction loss, the model is facilitated to learn a better temporal matching ability, which is essential in matching-based video tasks.

Before introducing the proposed ASAD, we firstly revisit the multi-head self-attention in ViT \cite{ViT}. Let $\bz \in \real^{N\times D}$ be the input sequence of the two concatenated input frames, $N$ denotes the total patch number in the two frames and $D$ is the feature dimension. The standard multi-head self-attention \cite{ViT} is:
\begin{align} 
[\bq, \bk, \bv] &= \bz \bU_{qkv}, 
\quad 
\label{softmax}
\text{SA}(\bz) = \text{softmax}(\tfrac{1}{\sqrt{D_{k}}}\bq \bk^{T}), \\
\text{MSA}(\bz) &= [\text{SA}_{1}(\bz); \text{SA}_{2}(\bz); \cdot \cdot \cdot; \text{SA}_{k}(\bz)]\bU_{m},
\end{align}
where  $\bU_{qkv} \in \real ^{D \times 3D_{k}}$ and $\bU_{m} \in \real ^{k\cdot D_{k} \times D}$. Let $A=\frac{\bq \bk^{T}}{\sqrt{D_{k}}} \in \real^{N\times N}$ denote the \emph{attention matrix}. Our ASAD performs spatial-attention dropout on $A$ {to remove some within-frame interactions}.

\begin{figure}
\begin{center}
   \includegraphics[width=1.0\linewidth]{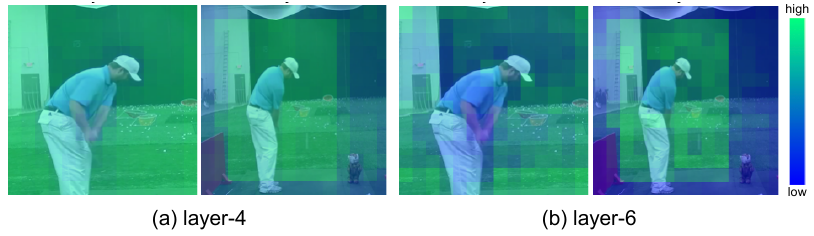} 
\end{center}
\vspace{-0.55cm}
 \caption{{Visualization of the temporal matching function $f_{tem}$ on an example frame pair. 
 }
 {A large value of $f_{tem}(i)$ indicates that the $i$-th pixel matches well to a pixel in the other frame.}
 }
\label{f_function}
\end{figure}

\noindent\textbf{Temporal matching probability.} 
{We first need to consider the best tokens on which to apply ASAD. 
Intuitively, a query token that has a strong match in the other frame should be a good candidate, since, in the absence of within-frame cues, it can still be reconstructed well using the temporal cues in the other frame.}
Here, we define a temporal matching function $f_{tem}(\cdot)$ to measure the temporal matching probability of the $i$-th query token:
\begin{align}
f_{tem}(i) = \max_{j \in \Omega_{t}(i)} (\hat{A}_{i,j}), \quad
\hat{A} = \text{softmax}_{\text{row}}(A),
\end{align}
where the $\text{softmax}$ function is applied on each row of $A$, 
$f_{tem}(i) \in [0,1]$, and $\Omega_{t}(i)$ denotes the \emph{temporal} index set of the $i$-th query token, which contains all the token indices of the other frame. A larger value of $f_{tem}(i)$ indicates a larger probability that the $i$-th query token is well-matched in the other frame, {and thus a good candidate for ASAD}. {A visualization of $f_{tem}(\cdot)$ is  in Fig.~\ref{f_function}.}

\noindent\textbf{Overall dropout probability measurement.} 
The overall spatial-attention dropout probability at location $(i,j)$ is measured by using both the temporal matching probability and the normalized spatial importance:
\begin{align} 
W_{i,j} = f_{tem}(i)  \tfrac{\hat{A}_{i,j}}{\sum_{j'\in \Omega_{s}(i)}{\hat{A}_{i,j'}}}, \quad
j\in\Omega_s(i), 
\end{align}
where $\Omega_{s}(i)$ is the \emph{spatial} index set that contains all the other token indices (i.e., excluding the query index itself) in the same frame as the $i$-th query. When  $W_{i,j}$ is large,  the $i$-th query token has a good \emph{between-frame} match, and meanwhile the $j$-th \emph{within-frame} token is important for the $i$-th query. In this case, dropping the within-frame attention element $(i,j)$ in $A$ facilitates the model to use {between-frame} (temporally-matched) tokens for token learning or reconstruction. {Finally, we set the dropout probability for self-attention and temporal-self-attention to be 0, i.e., $W_{i,i}=W_{i,i+N/2}=0$.}

Note that there are $N(N/2-1)$ (i.e., excluding self-attention elements) spatial-attention elements in total, and only these spatial-attention elements are considered for dropout.
With a pre-defined dropout ratio $P$, we globally drop 
a total  of $N_{d} = P  N(N/2-1)$ attention elements from $A$.


\noindent \textbf{Sampling for Dropout.} We draw $N_{d}$ elements from a multinomial distribution based on the dropout probability matrix $W$. Then we drop the elements in $A$ with the corresponding indices by setting their values to $-\infty$. After applying the softmax function in (\ref{softmax}), the corresponding spatial-attention weights are removed. The other operations are the same with the original multi-head self attention mechanisms used in ViT. 
The PyTorch-like pseudocode is presented in Algorithm \ref{asad_algorithm}.

\begin{algorithm}[t]
\footnotesize
\SetAlgoLined
    \PyComment{Input: attention matrix A, sequence length N, drop number $N_d$} \\
    \PyCode{W = torch.zeros\_like(A)} \PyComment{N-by-N}  \\
    \PyCode{A = A.detach().{\color{red}{softmax}}(dim=-1))} \PyComment{N-by-N}  \\
    \PyCode{}   \\
    \PyComment{get temporal attentions in each row of A}  \\
     \PyCode{A\_tem = temporal\_index(A)} \PyComment{N-by-N//2}  \\
     \PyCode{f\_tem = A\_tem.{\color{red}{max}}(dim=-1).values} \PyComment{N-by-1}  \\
     \PyCode{}   \\
     \PyComment{get spatial attentions in each row of A}  \\
     \PyCode{A\_spa = spatial\_index(A)} \PyComment{N-by-N//2}  \\
     \PyComment{avoid self-attention dropout}  \\
     \PyCode{A\_spa[0:N//2, 0:N//2].{\color{red}{fill\_diagonal\_}}(0)}  \\
     \PyCode{A\_spa[N//2:, 0:N//2].{\color{red}{fill\_diagonal\_}}(0)}  \\
     \PyCode{A\_spa=A\_spa/A\_spa.{\color{red}{sum}}(dim=-1, keepdim=True)} \\
     \PyCode{}   \\
     \PyComment{calculate overall dropout probability}  \\
      \PyCode{f\_all = f\_tem * A\_spa}  \PyComment{N-by-N//2}  \\
      \PyCode{}   \\
      \PyComment{put back to probability matrix W}  \\
       \PyCode{W[0:N//2, 0:N//2] = f\_all[0:N//2, 0:N//2]} \\
       \PyCode{W[N//2:, N//2:] = f\_all[N//2:, 0:N//2]} \\
       \PyComment{sample $N_d$ elements based on W}  \\
       \PyCode{}   \\
       \PyCode{indices={\color{red}{torch.multinomial}}(W.view(1,-1),$N_d$)}\\
        \PyCode{{\color{red}{return}} indices} \\
       
\caption{ASAD Pseudocode, PyTorch-like}
\label{asad_algorithm}
\end{algorithm}

\noindent\textbf{Autoencoder and Training.} 
Our ASAD method 
has negligible additional time cost compared with TwinMAE, due to the efficient matrix operation in GPUs. We apply ASAD to each layer in the decoder during the pre-training stage, {so as to learn encoder representations that support temporal matching}. 
\jimmyy{As demonstrated in Fig. \ref{temp_spatial_attention_comp}, our DropMAE with ASAD leverages more \emph{between-frame} attentions for reconstruction, which learns to perform accurate temporal matching in order to recover the patches between frames, thus leading to better temporal correspondence learning.}

In the next section, we introduce downstream task fine-tuning based on the pre-trained ViT model.

\subsection{Downstream Temporal Matching Tasks}
\label{down_temp_match}
After obtaining the pre-trained DropMAE model, we fine-tune the well-learned encoder (i.e., the ViT model) on downstream temporal matching tasks. To demonstrate the generality of DropMAE's learned representations, here we consider 6 downstream tasks: video object tracking, video object segmentation, 3d point cloud tracking, optical flow estimation, long-term point tracking, and self-supervised visual correspondence learning.

\subsubsection{Video Object Tracking}
Recently, the MAE ViT models {pre-trained on ImageNet} are applied to VOT and show impressive results. We use the representative tracker OSTrack \cite{ostrack} as our baseline tracker for fine-tuning. In OSTrack, the cropped template and search images are firstly serialized into sequences and concatenated together. Then the overall sequence is added with the positional embeddings and input to the ViT backbone for joint feature extraction and interaction. Finally, the updated search features are input to a prediction head to predict the target bounding box. 

During the fine-tuning stage, we use our pre-trained DropMAE {encoder} weights to initialize the ViT backbone used in OSTrack. Meanwhile, to keep consistency with the pre-training stage, two {frame identity }\CUT{temporal positional }embeddings are respectively added to template and search embeddings. {We use the same training losses of the original OSTrack. We denote this DropMAE-based tracker as DropTrack. 



\subsubsection{Video Object Segmentation}
For VOS, there are currently no methods based on ViT. Thus, we build a simple VOS baseline with a ViT backbone, namely DropSeg, to bridge this gap. The overall pipeline of our DropSeg is shown in Fig. \ref{vos_pipeline}.

\noindent\textbf{Input serialization.} Given a template frame with a binary mask, VOS aims to segment the object-of-interest in each frame of a video. {Similar to the pre-training stage, the binary mask map, template and search frames are firstly converted to patch sequences, and then linearly projected and added with positional embeddings. Two {frame identity embeddings} are added to the template and search embeddings, and the mask embeddings are added to the template embeddings for mask encoding.}

  \begin{figure}
  \vspace{-0.2cm}
\begin{center}
   \includegraphics[width=0.85\linewidth]{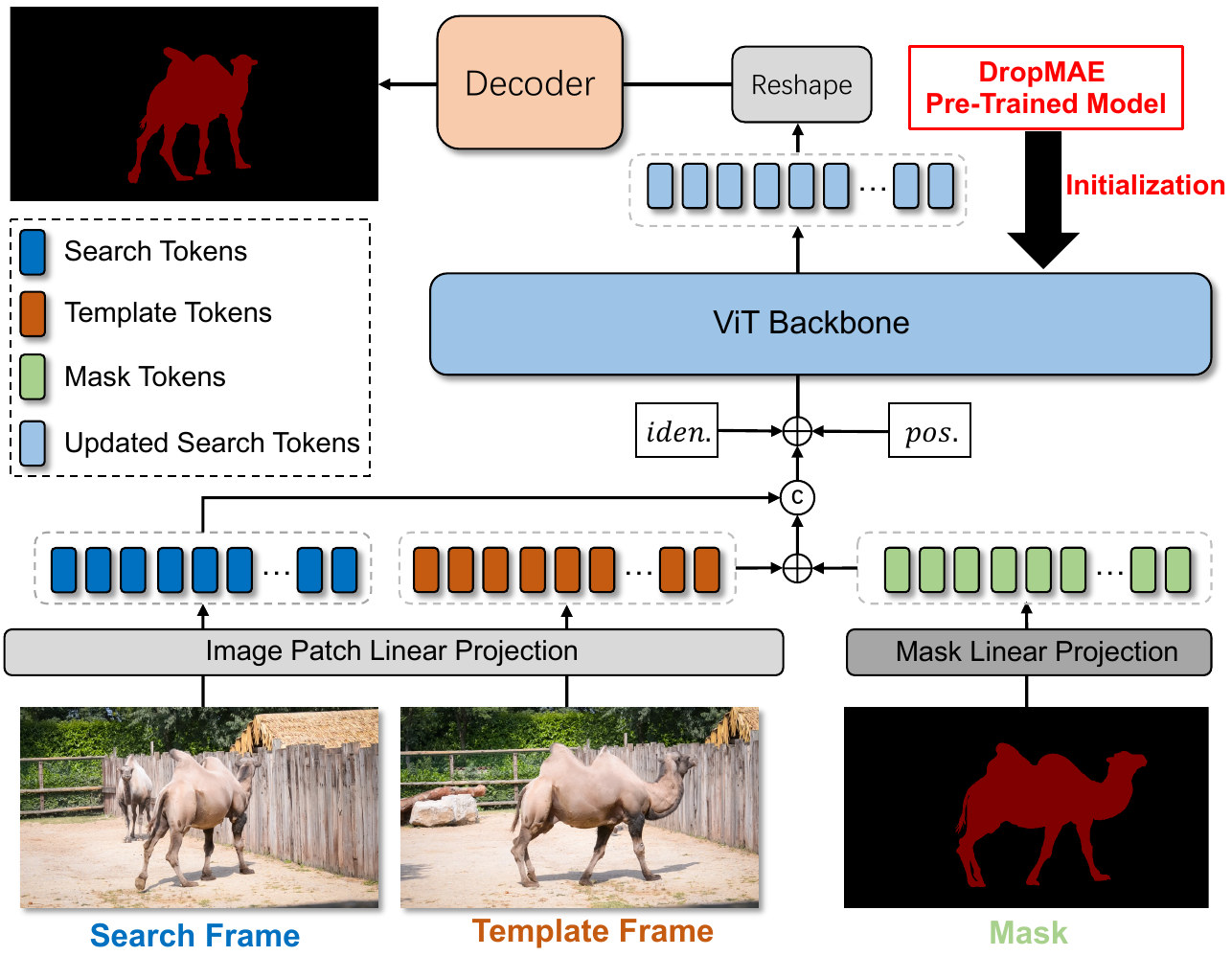} 
\end{center}
\vspace{-0.5cm}
 \caption{The overall pipeline of the proposed DropSeg for VOS.  \emph{iden} and \emph{pos} indicate frame identity embeddings and positional embeddings.}
\label{vos_pipeline}
\end{figure}


\noindent\textbf{Joint feature extraction and interaction.} The template and search embeddings are concatenated together and input to the ViT backbone for joint feature extraction and matching. We use the updated search features extracted from the last layer of ViT for mask prediction.

\noindent\textbf{Mask prediction.} The existing VOS approaches \cite{STM,STCN,XMEM,AOT} employ multi-resolution features for mask prediction. However, the updated search features are single-resolution. We follow \cite{plaindet} to upsample the search features to $2\times$ and $4\times$ sizes via two deconvolutional modules.  Finally, we use the same decoder used in \cite{STCN,STM} for mask prediction.

\noindent\textbf{Training loss.} We use the commonly-used cross entropy loss \cite{STCN,STM} to train the whole network architecture.

\noindent\textbf{Online inference.} During the online inference, we use the first frame with the mask annotation as the memory frame for online target matching in the search frame.  


\subsubsection{3D Point Cloud Tracking}
Given the 3D bounding box  of a target in the first frame, 3D single object point cloud tracking aims to estimate the target's bounding boxes in the following frames. The key success of this task is to perform robust matching between the target point cloud and search point could data. 
Here we investigate the effectiveness of our DropMAE in 3D matching learning. Specifically, we mainly follow SiamDisst \cite{wu2023boosting}, which employs a ViT as the backbone for joint feature extraction and matching, and a 3D bounding box head for box regression. We initialize the ViT backbone in the 3D tracker SaimDisst with our DropMAE pre-trained weights for more robust template matching. 
Interestingly, we find that DropMAE pre-trained model is also effective for 3D point tracking, even though the model is pre-trained only on 2D videos. 
We hope that our research could inspire more research about transferring 2D pre-trained models to 3D tracking tasks.

\subsubsection{Optical Flow Estimation}
Optical flow focuses on short-term dense motion estimation between consecutive frames. RAFT \cite{raft} is a typical optical flow approach, which predicts a field of pixel-wise 2D vectors through iterative refinement. Specifically, RAFT \cite{raft} predicts a dense 2D flow field between two adjacent RGB frames in three main steps: (1) using feature and context encoders to extract low-resolution feature maps from input images; 2) constructing a full correlation volume map by computing visual similarity between feature map pairs in the two input frames; 3) iteratively refining the flow predictions with an RNN unit. 

In this paper, we use RAFT as our baseline to incorporate our DropMAE pre-trained model for better temporal matching. In RAFT, feature extraction and matching are performed separately. RAFT  uses a ResNet-based feature encoder \cite{resnet} to extract the features of the input frames firstly, and then employs a manually-designed matching layer (i.e., implemented as matrix multiplication) to calculate the visual similarity maps. However, this separate scheme cannot extract dynamic target-aware features for better visual similarity calculation. \jimmyy{To address this problem, we can equip RAFT with a ViT feature extractor for target-aware feature extraction. However, in preliminary experiments, naively employing the ViT with some off-the-shell pre-training weights (e.g., random, supervised ImageNet and MAE \cite{mae}) for RAFT causes significantly performance degradation (see Table \ref{comp_optical_flow}), which is mainly due to the lack of rich  temporal prior in these pre-training weights.}

To alleviate this, we use our DropMAE pre-trained model as the feature encoder,  \jimmyy{enabling RAFT to effectively use a ViT-based feature extractor.} The new tracker (denoted as DropRAFT) enables RAFT to perform joint feature extraction and matching between the two input frames, which produces a more accurate correlation volume and significantly enhances the final performance (see Table \ref{comp_optical_flow}). The whole process can be written as:
\begin{align}
\label{eq::optical}
I'_{1}, I'_{2} = E_{D}(I_{1},I_{2}), \quad \quad \quad \quad \quad \\
C_{k} = \text{AvgPool}(I'_{1} \cdot {I'_{2}}^{T}, 2^{k}) \in \real^{H \times W \times \frac{H}{2^{k}} \times \frac{W}{2^{k}}},
\end{align}
where $E_{D}(\cdot,\cdot)$ is our pre-trained DropMAE encoder, which takes the concatenated patches of the frames $I_{1}$ and $I_{2}$ as the input for joint feature extraction and interaction, obtaining $I' \in \real^{H \times W \times D}$. $C_{k}$ is the $k$-th correlation volume, which is obtained by using average pooling with kernel sizes $\{2^{k}\}_{k=0}^{3}$ on the last two dimensions of $I'_{1} \cdot {I'_{2}}^{T} \in \real^{H \times W \times H \times W}$. After obtaining $\{C_{k}\}_{k=0}^{3}$, we use the same iterative refinement in RAFT to obtain the flow predictions. In this work, we show that a more accurate correlation volume $I'_{1} \cdot {I'_{2}}^{T}$ obtained by our DropMAE can significantly improve the optical flow estimation accuracy.

  \begin{figure}
\begin{center}
   \includegraphics[width=1.0\linewidth]{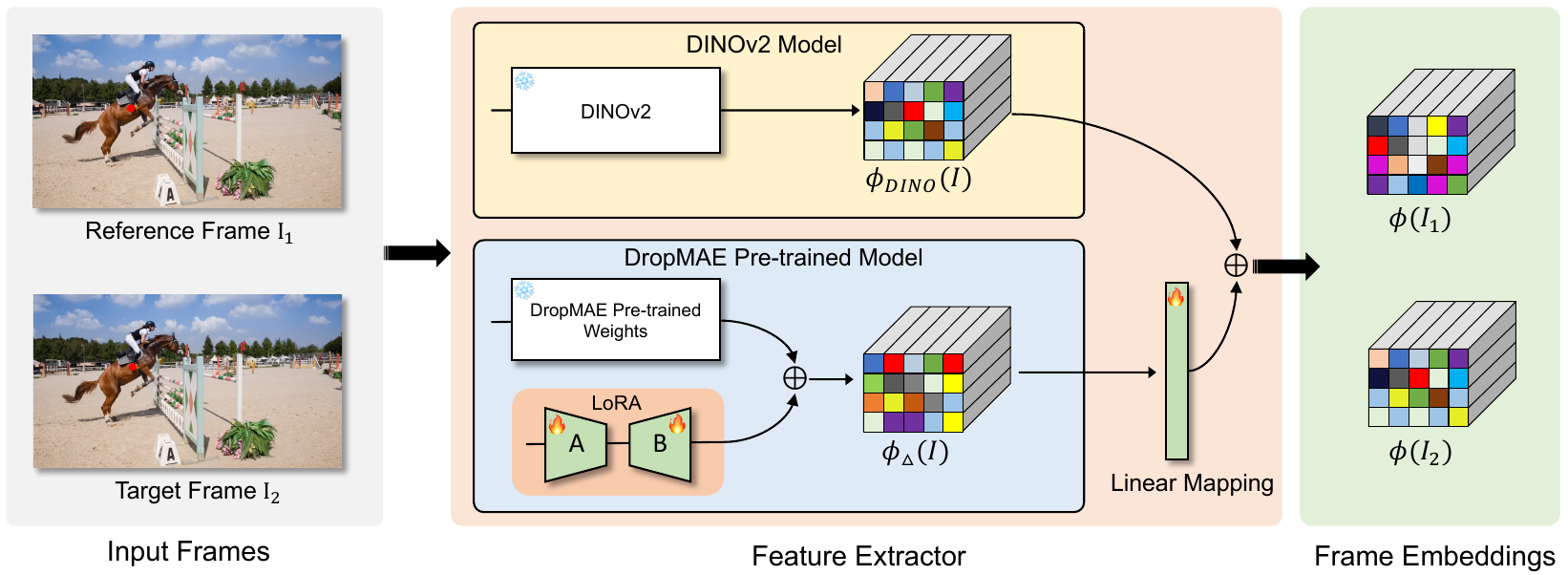} 
\end{center}
\vspace{-0.55cm}
 \caption{Overall pipeline of the proposed DropDINO using parameter-efficient LoRA training with our DropMAE pre-trained model for long-term point tracking.}
\label{dropdino}
\end{figure}


\subsubsection{Long-Term Point Tracking}\label{sec:point_track}
Long-term point tracking focuses on matching corresponding points across distant frames in a video. TAP-Vid \cite{doersch2022tap} proposes to solve this problem by formulating it as \emph{tracking any point} (TAP) and using a CNN baseline w/ a synthetic dataset. Recently, Dino-Tracker \cite{tumanyan2025dino} combines the online learned residual CNN model with a pre-trained DINO-ViT model \cite{oquab2023dinov2}, achieving state-of-the-art long-term point tracking performance.

\noindent\textbf{Revisit of Dino-Tracker.} Given a query point in an initial video frame, TAP aims to track the query points in the subsequent frames, accurately estimating its trajectories and occlusion status in the long-term. To adapt to a specific online tracking video, Dino-Tracker online optimizes a residual CNN model with the combination of a pre-trained DINO-ViT model. The key idea of Dino-Tracker is to predict residuals to the pre-trained DINO-ViT model via the residual CNN model. The residual representations are supposed to be effective in capturing temporal correspodences, which are complementary to the semantic representations in DINO features. The feature combination can be formulated as: 
\begin{align}
\label{eq::dinotrack}
\phi(I) = \phi_{DINO}(I) + \phi_{\Delta}(I),
\end{align}
where $\phi_{DINO}(I)$ and $\phi_{\Delta}(I)$ are respectively DINO and residual features.

\noindent\textbf{Limitations.} DINO-Tracker implements $\phi_{\Delta}(\cdot)$ as a CNN model (i.e., ResNet), and it  claims that the CNN model can effectively benefit from its inductive bias and encode similar RGB patches across frames into similar feature representation. Moreover, $\phi_{\Delta}(\cdot)$ is zero initialized for each online testing video. However, CNN-based $\phi_{\Delta}(\cdot)$ has the following limitations: 1) the limited expressive power of CNN may cause inaccurate matching to similar or distractor points; 2) zero-initialized $\phi_{\Delta}(\cdot)$ may not be optimal  for the online test video adaptation with limited training data. To address the aforementioned issues, we propose to use our DropMAE pre-trained model as $\phi_{\Delta}(\cdot)$ for online temporal matching learning. We call this new tracker as DropDINO. We find that: 1) DropMAE is a strong temporal learner in the low data regime of online adaptation, which  well complements the DINO features used in Dino-Tracker; 2) fine-tuning fewer parameters than the CNN based $\phi_{\Delta}(\cdot)$ achieves better tracking performance, demonstrating the effectiveness of DropMAE.

\begin{figure}
\begin{center}
   \includegraphics[width=1.0\linewidth]{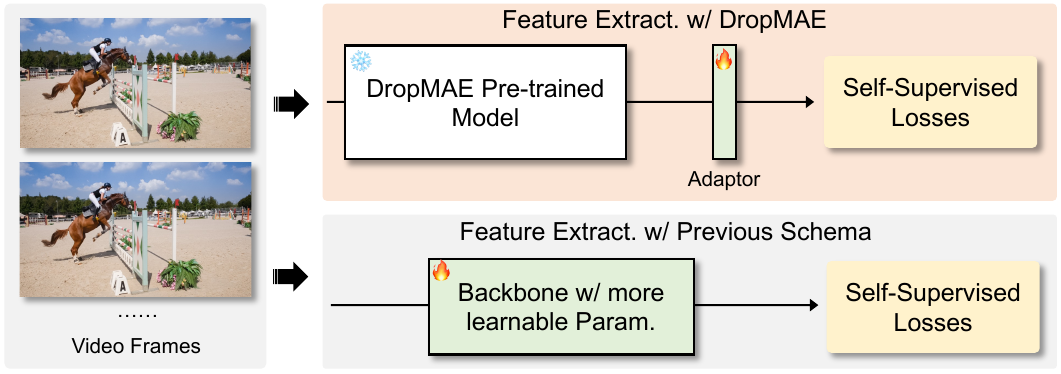} 
\end{center}
\vspace{-0.55cm}
 \caption{Overall pipeline of parameter-efficient self-supervised representation learning with our DropMAE pre-trained model.}
\label{self_drop}
\end{figure}


\noindent\textbf{Parameter-Efficient LoRA Training.} Since DropMAE adopts a ViT-base model, which has higher model complexity than the CNN based $\phi_{\Delta}(\cdot)$ in the original Dino-Tracker, we adopt LoRA training \cite{hu2022lora} to enable parameter-efficient training, which is illustrated in Fig. \ref{dropdino}. Note that \cite{tumanyan2025dino} attempted to implement $\phi_{\Delta}(\cdot)$ as the DINOv2 ViT model with the LoRA fine-tuning. However, the performance was significantly degraded, which is mainly because the rich semantic features in DINOv2 VIT are not suitable for temporal fine-tuning. In contrast, due to the well-learned temporal prior, our DropMAE achieves SOTA performance by applying parameter-efficient LoRA training.

\begin{table*}[t]
  \newcommand{\tabincell}[2]
  \centering
    \footnotesize \centering
       \caption{{Comparison of pre-training methods on} downstream VOT and VOS tasks on GOT-10k \cite{got10k} and DAVIS-17 \cite{davis17}. All methods adopt the ViT-B/16 model \cite{ViT} with 224$\times$224 input images for pre-training. The pre-training time is measured on 64 NVIDIA V100 GPUs.
   The best two results are shown in \textcolor{red}{\textbf{red}} and \textcolor{blue}{\textbf{blue}}.} 
   \vspace{-0.3cm}
    \begin{tabular}{cccc|ccc|ccc}
    \Xhline{\arrayrulewidth}
    \multirow{2}{*}{Methods}&  
  \multirow{2}{*}{Pre-training Data}& \multirow{2}{*}{Epochs}&\multirow{2}{*}{Pre-train. Time (h)}&\multicolumn{3}{c|}{GOT-10k (VOT)} & \multicolumn{3}{c}{DAVIS-17 (VOS)} \cr
  & & & & AO & SR$_{0.5}$ & SR$_{0.75}$ & 
  $\mathcal{J}\&\mathcal{F}$ & $\mathcal{J}$ & $\mathcal{F}$ 
  \cr
     \Xhline{\arrayrulewidth}  
     No Pre-training              &-   &- &- &62.7&72.8&53.7 &69.5&66.9&72.2      \cr
     Supervised IN1k \cite{deit}              &IN1K   &300 &-&69.7&79.0&65.6  & 78.0&74.8&81.1     \cr
     Supervised IN21k  \cite{imagenet21k_learn}            &IN21K  &80 &-&70.2&80.7&65.4 &78.5&75.4&81.7      \cr
       CLIP \cite{clip}              &IN1K   &32 &- &67.4	&76.8 &60.0 &73.6&70.5&76.7      \cr  
        MOCO-v3 \cite{mocov3}              &IN1K   & 300 &-  & 70.1 & 80.1 & 65.3 & 78.4&75.4&81.5  \cr  
        BeiT \cite{beit} 		&IN1K   & 800 &103.1  & 67.4 & 76.8 &	60.0 &76.1&72.7&79.4    \cr  
        MAE \cite{mae} 		&IN1K   &1600 &84 &  73.7&83.2& \textcolor{blue}{\textbf{70.8}} & 81.7&78.5&84.9      \cr  
        VideoMAE \cite{videomae} &K400 & 1600 & 123.4 & 61.6 & 72.7 & 48.4 & - & - & - \cr
        \Xhline{\arrayrulewidth}  
        TwinMAE  &K400   &400 &20.7&72.2&83.2&65.9 & 79.3&76.4&82.3       \cr  
        TwinMAE  &K400   &800 &41.3&72.9&83.6&68.5 & 80.7&77.9&83.6     \cr  
        TwinMAE  &K400   &1600 &82.7&74.2&84.9&69.4  & 81.2&78.1&84.2     \cr  
     \Xhline{\arrayrulewidth}  
     \textbf{DropMAE} &K400	  &400 &21.1&73.2&83.9&67.5 & 81.3&78.5&84.0     \cr    
      \textbf{DropMAE} &K400	  &800 &42.2&74.8&85.4&70.5 & 82.7&\textcolor{blue}{\textbf{79.7}}&85.6     \cr    
       \textbf{DropMAE} &K400	  &1600 &84.4 &\textcolor{blue}{\textbf{75.8}}&\textcolor{blue}{\textbf{86.4}}&\textcolor{red}{\textbf{72.0}} & \textcolor{red}{\textbf{83.1}}&\textcolor{red}{\textbf{80.2}}&\textcolor{red}{\textbf{86.0}}     \cr    
       \textbf{DropMAE} &K700	  &800 &92.4 & \textcolor{red}{\textbf{75.9}}&   \textcolor{red}{\textbf{86.8}}&  \textcolor{red}{\textbf{72.0}} & \textcolor{blue}{\textbf{83.0}}&\textcolor{red}{\textbf{80.2}}&\textcolor{blue}{\textbf{85.7}}    \cr    
   \Xhline{\arrayrulewidth}  
   \end{tabular}  
  \label{pretrain_comp}
\end{table*}







\subsubsection{Self-Supervised Visual Correspondence Learning}
Generative MAE pre-training and its subsequent developments mainly focus on learning representations for downstream task fine-tuning. In the previous subsections, we considered the effectiveness of our DropMAE pre-training in fine-tuning based downstream temporal matching tasks. For completeness, we further evaluate the raw DropMAE representations in unsupervised tracking tasks, without any supervised fine-tuning. 
Here we also show that DropMAE  serves as effective pre-trained weights in the existing self-supervised learning method, DUL \cite{DUL}. Notably, leveraging DropMAE as pre-trained weights in \cite{DUL} results in significant speedup for self-supervised visual correspondence learning in videos.

\noindent\textbf{{Computing Affinity Matrix via DropMAE.}}
\jimmyy{The affinity matrix represents the similarity among visual features extracted from two consecutive frames in a video. Currently, self-supervised learning approaches use probabilities from the affinity matrix to find temporal correspondences. Specifically,} 
given a pair of consecutive video frames $I_{t}$ and $I_{t+1}$, we first use our DropMAE pre-trained model to extract frame features, obtaining $I'_{t}$ and $I'_{t+1} \in \real^{H \times W \times D}$. For the $i$-th query token feature $\mathbf{q}_{t}^{i}$ in the $t$-th frame, its affinity $\hat{k}_{t}^{t+1}(i,j)$ to the $j$-th feature in the $(t+1)$-th frame can be calculated as:
\begin{align}
\label{eq::affinity}
\hat{k}_{t}^{t+1}(i,j) = \text{Softmax}(I'_{t}, I'_{t+1}, \tau)_{i,j} = \tfrac{\text{exp}(\mathbf{q}_{t}^{i} \odot \mathbf{q}_{t+1}^{j} / \tau)}{\sum_{s=1}^{N}\text{exp}(\mathbf{q}_{t}^{i} \odot \mathbf{q}_{t+1}^{s} / \tau)}, 
\end{align}
where $\hat{K}_{t}^{t+1}=[\hat{k}_{i,j}]_{N\times N}$ is the frame affinity matrix, $N$ is the number of spatial features in a frame, $\odot$ indicates the inner product and $\tau$ is the temperature hyper-parameter.

\noindent\textbf{Label Propagating via $\hat{K}_{t}^{t+1}$.} Each row in $\hat{K}_{t}^{t+1}$ indicates the correlation between a query feature in the last frame and the tokens in the current frame. For fair comparison of learned representations, we follow \cite{crw,vfs} to propagate different types of labels in the previous frame to the current frame. Specifically, given mask labels $M \times \real^{N\times 1}$ in the $t$-th frame, its predicted masks in the $(t+1)$-th frame can be obtained by $K_{t}^{t+1}M$. Mask propagation follows a recurrent process, where the output mask from the current frame serves as input for the subsequent frame. As in \cite{crw,vfs}, we only keep the top 10 values for each row and set other values to zero. For box and pose label propagation, following \cite{unitrack} and \cite{vfs}, we use the Gaussian belief maps and SiamFC for unsupervised pose and object tracking.

\noindent\textbf{Self-supervised Learning w/ DropMAE.} 
Contrastive learning methods \cite{DUL} have demonstrated strong performance in self-supervised correspondence learning. Interestingly, we find that our DropMAE  pre-trained model can be effectively integrated with existing contrastive learning frameworks for efficient self-supervised correspondence learning. As illustrated in Fig. \ref{self_drop}, we adopt DUL \cite{DUL} as our baseline and replace its feature extractor with the DropMAE pre-trained model. Additionally, we add a lightweight adaptor  on top of the DropMAE features, training it using the self-training and cross-view consistency losses from \cite{DUL} while keeping the DropMAE backbone frozen. Notably, our DropMAE variant achieves a 16.6× self-supervised learning speedup on K400 while learning more effective representations than the baseline \cite{DUL} with fewer parameters, enabling efficient self-supervised correspondence learning.

\newcommand{\cmark}{\ding{51}}%
\newcommand{\xmark}{}%
\begin{table*}[t]
\vspace{-0.2cm}
\centering
\footnotesize
\caption{Comparison with state-of-the-art VOT approaches on four large-scale  datasets.
The best two results are shown
in \textcolor{red}{\textbf{red}} and \textcolor{blue}{\textbf{blue}}. For GOT-10k evaluation, all the methods follow the one-shot protocol, training only on the training set in GOT-10k. Our DropTrack-B384 achieves SOTA performance w/o using complex  temporal updating (TU).}
\vspace{-0.3cm}
\centering
\begin{tabular}{c|c|c|ccc|cc|ccc|ccc}
\hline
 \multirow{2}{*}{\begin{tabular}[c]{@{}c@{}}Method\end{tabular}} & \multirow{2}{*}{\begin{tabular}[c]{@{}c@{}}TU\end{tabular}}&\multirow{2}{*}{\begin{tabular}[c]{@{}c@{}}Source\end{tabular}}& \multicolumn{3}{c|}{GOT-10k \cite{got10k}}& \multicolumn{2}{c|}{TNL2K \cite{tnl2k}}  & \multicolumn{3}{c|}{LaSOT$_{\text{ext}}$ \cite{lasotext}}  & \multicolumn{3}{c}{LaSOT \cite{lasot}}                                                  \\
                            &                               &                                                                                           & AO           & SR$_{0.5}$           & SR$_{0.75}$           & AUC       & P          & AUC           & P$_{Norm}$         & P       & AUC           & P$_{Norm}$         & P         \\

\hline
    MDNet \cite{MDNet}      &\cmark & CVPR16                                                                                 & 29.9 & 30.3 & 9.9   & - & - & 27.9 & 34.9 & 31.8 & 39.7 & 46.0 & 37.3               \\
     ECO \cite{ECO}                 &\cmark & ICCV17                                                                          & 31.6 & 30.9 & 11.1          & 32.6 & 31.7 & 22.0 & 25.2 & 24.0 & 32.4 & 33.8 & 30.1             \\
           DiMP \cite{dimp}             &\cmark & ICCV19       & 61.1 & 71.7 & 49.2          & 44.7 & 43.4  & 39.2 & 47.6 & 45.1 &56.9 & 65.0 & 56.7           \\
           SiamR-CNN \cite{siamrcnn} &\cmark & CVPR20       & 64.9 & 72.8 & 59.7          &  52.3 & 52.8          & -         & -             & -             &64.8 & 72.2 & -              \\
           LTMU \cite{ltmu}       &\cmark & CVPR20                                                                                      & -          & -          & -          & 48.5          & 47.3          & 41.4 & 49.9 & 47.3 & 57.2 & -  &57.2             \\
        Ocean \cite{ocean}   &\cmark & ECCV20                                                                          & 61.1 & 72.1 & 47.3 &38.4 & 37.7 & - & - & - & 56.0 & 65.1 & 56.6  \\ 
  	TrDiMP \cite{trdimp} &\cmark & CVPR21                                                                                 & 67.1 & 77.7& 58.3         & -          & -          & -          & -          & -          & 63.9 & - & 61.4           \\
	AutoMatch \cite{automatch}      &\cmark & ICCV21                                                                            & 65.2 & 76.6 & 54.3            & 47.2 & 43.5             & 37.6 & - &43.0             & 58.3 & - & 59.9               \\
           STARK \cite{stark}    &\cmark & ICCV21                                                                                & 68.8 & 78.1 & 64.1         & -          & -          & -             & -             & -             & 67.1 & 77.0            &   -             \\
            KeepTrack \cite{keeptrack}   &\cmark & ICCV21                                                                                    & -          & -          & -          & -          & -          & 48.2          & -          & -          & 67.1 & 77.2 & 70.2          \\
              MixFormer-L \cite{MixFormer}			&\cmark & CVPR22                  & 70.7 & 80.0& 67.8  & -          & -          & -          & -          & -          & 70.1 & 79.9 & 76.3           \\
              UAST \cite{uast}			& \cmark & ICML22                  & 63.5 & 74.1 & 51.4 & -          & -          & -         & -          & -          & 57.1 & - & 58.7           \\
              AiATrack \cite{AiATrack}                                                                &\cmark & ECCV22                                                                                      & 69.6 & 80.0 & 63.2  & -          & -          & 46.8 & 54.4 & 54.2          & 49.6 & 56.9 & 49.1          \\
                 CIA50 \cite{CIA50}			&\cmark & ECCV22                                                                                      & 67.9          & 79.0          &60.3 & 50.9          & \textcolor{blue}{\textbf{57.6}}          & -          & -          & -          & 66.2          & - & 69.6           \\
                  MixFormer-22k \cite{MixFormer} &\cmark & CVPR22 & 70.7 & 80.0 & 67.8 & - & - & - & - & - & 69.2 & 78.7 & 74.7 \\
                 SeqTrack-B384 \cite{seqtrack} &\cmark & CVPR23 & 74.5 & 84.3 & 71.4 & 56.4 & - & 50.5 & 61.6 & 57.5 & 71.5 & 81.1 & 77.8 \\
                 ARTrack-B384 \cite{artrack} &\cmark & CVPR23 & \textcolor{blue}{\textbf{75.5}} & 84.3 & \textcolor{red}{\textbf{74.3}} & - & - & \textcolor{blue}{\textbf{51.9}} & \textcolor{blue}{\textbf{62.0}} & \textcolor{blue}{\textbf{58.5}} & \textcolor{red}{\textbf{72.6}} & \textcolor{blue}{\textbf{81.7}} & \textcolor{red}{\textbf{79.1}} \\   
                      \Xhline{\arrayrulewidth}  
                    SiamFC \cite{SiamFC}     &\xmark &    ECCVW16                                                                                & 34.8 & 35.3 & 9.8  & 29.5 & 28.6 & 23.0 & 31.1 & 26.9 & 33.6 & 42.0 & 33.9  \\        
      SiamPRN++ \cite{SiamRPN_plus}    &\xmark & CVPR19        & 51.7 & 61.6 & 32.5          & 41.3 & 41.2 & 34.0 &41.6 & 39.6 & 49.6 & 56.9 & 49.1         \\
          TransT \cite{transt}  &\xmark & CVPR21          & 67.1 & 76.8 & 60.9             & 50.7 & 51.7             & -  & -  & - & 64.9 & 73.8 & 69.0            \\
              SBT \cite{sbt}			&\xmark & CVPR22                  & 70.4 & 80.8& 64.7  & -          & -          & -         & -          & -          & 66.7 & - & 71.1           \\
              SwinTrack-384 \cite{swintrack}   &\xmark & NeurIPS22                & 72.4 & 80.5 & 67.8          & 55.9 &  57.1          & 49.1             & -             & 55.6             &  71.3            & -    & 76.5          \\
                  SimTrack-L \cite{simtrack}                                                                &\xmark  & ECCV22                                                                                    & 69.8 & 78.8          & 66.0         & 55.6          & 55.7          & -             & -             & -             & 70.5             & 79.7 & -              \\
                 OSTrack-384  \cite{ostrack}   &\xmark & ECCV22     & 73.7 & 83.2 & 70.8          &  55.9          & 56.7          &  50.5 &  61.3 &  57.6          & 71.1 &  81.1 &  77.6              \\
                 OneTracker \cite{hong2024onetracker} & \xmark & CVPR24 & - & - & - & 58.0 & 59.1 & - & - & - & 70.9 & 79.9 & 76.5 \\
                 DiffusionTrack-B256 (2) \cite{diffusiontrack} &\xmark & CVPR24 & 75.2 & \textcolor{blue}{\textbf{85.9}} & \textcolor{blue}{\textbf{72.0}} &  \textcolor{blue}{\textbf{56.5}} &  57.3 & - & - & - & 70.7 & 80.0 & 77.3\\  
        \textbf{DropTrack-B384}      &\xmark & \textbf{Ours}  &\textcolor{red}{\textbf{75.9}} &\textcolor{red}{\textbf{86.8}} &\textcolor{blue}{\textbf{72.0}}            & \textcolor{red}{\textbf{56.9}} & \textcolor{red}{\textbf{57.9}} & \textcolor{red}{\textbf{52.7}} & \textcolor{red}{\textbf{63.9}} & \textcolor{red}{\textbf{60.2}} & \textcolor{blue}{\textbf{71.8}} & \textcolor{red}{\textbf{81.8}} & \textcolor{blue}{\textbf{78.1}}  \\
\hline
\end{tabular} 
\label{overall_results}
\end{table*}

\section{Experiments on Pre-training}\label{sec:exp}

In this section, we conduct experiments comparing our DropMAE with other pre-training methods on the VOT and VOS tasks.

\subsection{Implementation Details}

\noindent\textbf{Pre-training.} In the pre-training stage, we explore various large-scale video data sources to pre-train our DropMAE model, including Kinetics-400 \cite{k400} (K400), Kinetics-600 \cite{k600} (K600), Kinetics-700 \cite{k700} (K700), Moments in Time \cite{moments} (MiT) and WebVid-2M \cite{webvid}. The detailed performance comparison using different pre-training datasets is shown in the ablation study in \S\ref{sec:abl}. We use the standard ViT-B/16 \cite{ViT} as our backbone for pre-training,  following the training settings used in the original MAE \cite{mae}. For the dropout ratio $P$, we set $P=0.1$ following the ablation study in Fig. \ref{abl_p}. The pre-training is conducted on 64 NVIDIA V100 GPUs. 
As illustrated in Table \ref{pretrain_comp}, the 1600-epoch pre-training takes about 84 hours on K400 \cite{k400}, and it can be further reduced to 58 hours by using 64 NVIDIA A100 GPUs.

\noindent\textbf{VOT.} We use the training splits of LaSOT \cite{lasot}, COCO \cite{coco}, TrackingNet \cite{trackingnet} and GOT-10k \cite{got10k} for training our DropMAE-based tracker (denoted as DropTrack). For the GOT-10k evaluation, we follow the one-shot evaluation and only fine-tune the model on the training split of GOT-10k. We use a base learning rate of 2.5e-4 while keeping the other parameters  same as OSTrack \cite{ostrack}. The inference speed of our DropTrack is the same as the baseline OSTrack, which
is 58.1 FPS measured on a single GPU.

\noindent\textbf{VOS.} We use Youtube-VOS \cite{youtubevos} and Davis \cite{davis17} datasets for fine-tuning following the standard convention \cite{STM,STCN}. We use the Adam optimizer with a learning rate of 2e-5 for optimziation. The model is trained with 210,000 iterations and the learning rate is decayed at 125,000 iterations. The fine-tuning is conducted on 8 A100 GPUs, and the whole training takes about 16 hours.

\begin{table}[t]
\vspace{-0.4cm}
  \caption{Comparison with state-of-the-art VOT approaches on OTB100 \cite{OTB100}, ITB \cite{itb} and TrackingNet \cite{trackingnet}. The best two results are shown in \textcolor{red}{\textbf{red}} and \textcolor{blue}{\textbf{blue}}.}  
  \vspace{-0.3cm}
  \newcommand{\tabincell}[2]
    \centering
  \footnotesize
    \begin{tabular}{c|c|c|cc}
    \Xhline{\arrayrulewidth}
    \multirow{2}{*}{Method}&  
  \multicolumn{1}{c|}{OTB100}& \multicolumn{1}{c}{ITB}&\multicolumn{2}{|c}{TrackingNet} \cr
   & AUC & AUC & AUC & P$_{Norm}$ \cr

       \Xhline{\arrayrulewidth}  
     SiamFC \cite{SiamFC}             &58.3   &44.1 &  57.1 & 66.3  \cr  
     Ocean \cite{ocean}             &68.4   &47.7 & - & -      \cr  
     ATOM \cite{atom}             &68.3   &47.2 & 70.3 & 77.1    \cr  
     DiMP \cite{dimp}             &53.7   &339 & 74.0 & 80.1    \cr  
     TransT \cite{transt}             &\textcolor{blue}{\textbf{69.5}}   &54.7 & 81.4 & 86.7      \cr 
     STARK \cite{stark}              &68.1   &57.6  &  82.0 & 86.9    \cr  
     OSTrack \cite{ostrack}              &-   &\textcolor{blue}{\textbf{64.8}} & \textcolor{blue}{\textbf{83.9}} & \textcolor{blue}{\textbf{88.5}}   \cr  
     \textbf{DropTrack}            & \textcolor{red}{\textbf{69.6}}   & \textcolor{red}{\textbf{65.0}} &\textcolor{red}{\textbf{84.1}} &\textcolor{red}{\textbf{88.9}}      \cr  
   \Xhline{\arrayrulewidth}  
   \end{tabular}  
   \centering
  \label{otb_like_compare}
\end{table}

\subsection{Comparison with Pre-Training Methods} 
In Table \ref{pretrain_comp}, we compare our DropMAE with existing pre-training methods on the downstream tasks of VOT and VOS.
{DropMAE and TwinMAE are pre-trained using videos (K400, K700), while MAE and other methods are pre-trained on ImageNet 1k or 21k (IN1K, IN21K).}
The VOT and VOS baselines illustrated in Sec. \ref{down_temp_match} use the official pre-trained VIT-B/16 models provided by existing pre-training approaches (see Table \ref{pretrain_comp}) for fine-tuning. 
TwinMAE with 800-epoch training performs favorably against MAE on VOT, but achieves inferior results on VOS. There are two main reasons: 1) TwinMAE is not effective enough at learning temporal matches; 2) The number of object classes in K400 is limited, and meanwhile the object classes in DAVIS-17 are included in ImageNet. Thus MAE generalizes well to VOS. Our DropMAE, which is a stronger temporal matching learner, outperforms MAE on both the VOT and VOS tasks with 800-epoch training (i.e., 42.2 hours) by using the K400 dataset. This indicates that our DropMAE is $2\times$ faster than MAE.

 \begin{figure}
 \vspace{-0.4cm}
\begin{center}
   \includegraphics[width=0.9\linewidth]{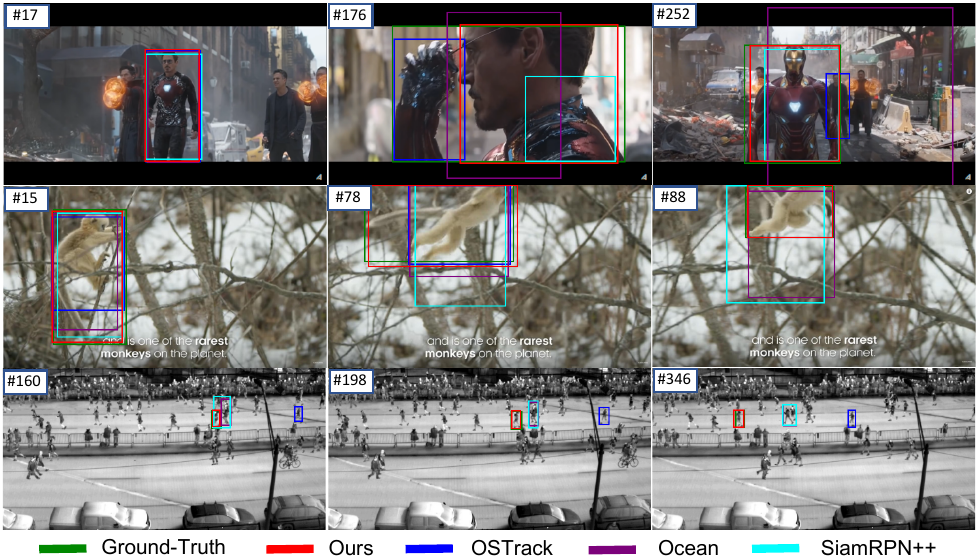} 
\end{center}
\vspace{-0.55cm}
 \caption{Qualitative VOT results of our DropTrack and several compared methods, including OSTrack \cite{ostrack}, Ocean \cite{ocean} and SiamRPN++ \cite{SiamRPN_plus}. The three video sequences are collected from TNL2K \cite{tnl2k}. The frame number is shown in the \emph{top-left} of each frame.} 
\label{vot_qualitative}
\end{figure}


\section{Experiments on Downstream Tasks}\label{sec:sota_comparison}
In this section, we compare our fine-tuned models for six downstream tasks with state-of-the-art approaches on various   benchmarks. 
We use DropMAE trained on K700 with 800 epochs as the pre-trained model for both VOT and VOS fine-tuning. For the other downstream  tasks, DropMAE trained on K400 with 1600 epochs is used as the pre-trained model for a fair comparison w/ other video-based self-supervised approaches.

\subsection{Implementation Details}

\noindent\textbf{3D Point Cloud Tracking.} Following \cite{wu2023boosting}, we use  the first 6 layers of ViT-Base model as our backbone for joint 3D point feature extraction and interaction. The backbone is initialized with our DropMAE pre-trained model for further fine-tuning on the Van category of the KITTI dataset \cite{geiger2013vision} with the same training hyper-parameters used in \cite{wu2023boosting}.

\noindent\textbf{Self-Supervised Correspondence Learning.} We replace the backbone in DUL \cite{DUL} with our DropMAE pre-trained model and add a learnable adaptor, implemented as a lightweight residual block (with 5.4M parameters, as detailed in Table \ref{corres_compex}) for representation learning. We adopt the same training settings from DUL, except that only one  training epoch is used due to our robust pre-trained weights.

\begin{table}[t]
 \caption{Comparison with state-of-the-art VOS approaches on the validation sets of DAVIS-2016 \cite{davis16} and DAVIS-2017 \cite{davis17}. OL, M and S indicate \textbf{O}nline \textbf{L}earning, using \textbf{M}emory mechanism, and using \textbf{S}ynthetic videos for pre-training.}
 \vspace{-0.3cm}
 \resizebox{\linewidth}{!}{
 \footnotesize
    \centering
    \begin{tabular}{@{}c@{\hspace{0.1cm}}|c@{\hspace{0.2cm}}c@{\hspace{0.2cm}}c|ccc|ccc@{}}
    \Xhline{\arrayrulewidth}
    \multirow{2}{*}{Method}& \multirow{2}{*}{OL}&
  \multirow{2}{*}{M} &
  \multirow{2}{*}{S}
  &\multicolumn{3}{c|}{DAVIS-2016 \cite{davis16}} & 
  \multicolumn{3}{c}{DAVIS-2017 \cite{davis17}} \cr
  &&&
  &\multicolumn{1}{c}{$\mathcal{J}\&\mathcal{F}$} &\multicolumn{1}{c}{$\mathcal{J}$} &\multicolumn{1}{c|}{$\mathcal{F}$} & \multicolumn{1}{c}{$\mathcal{J}\&\mathcal{F}$} &\multicolumn{1}{c}{$\mathcal{J}$} &\multicolumn{1}{c}{$\mathcal{F}$}  \cr
     \Xhline{\arrayrulewidth}  
     RANet   \cite{ranet}                 &\xmark & \xmark & \cmark &85.5& 85.5& 85.4& 65.7& 63.2& 68.2     \cr    
     STM \cite{STM}                 &\xmark & \cmark & \cmark   & 89.3 & \textcolor{blue}{\textbf{88.7}} & 89.9 & 81.8 & 79.2& 84.3   \cr 
     FRTM \cite{frtm}                 &\cmark & \cmark &\xmark &83.5 & 83.6 & 83.4 & 76.7 & 73.9 & 79.6      \cr    
     TVOS \cite{tvos}                &\xmark & \cmark & \xmark & - &-& -& 72.3& 69.9& 74.7     \cr  
     LWL \cite{LWL}                &\cmark & \cmark & \xmark &- &-& - &81.6& 79.1& 84.1      \cr 
     CFBI \cite{CFBI}                 &\xmark & \cmark & \xmark  &89.4 &88.3& \textcolor{blue}{\textbf{90.5}}& 81.9& 79.1& 84.6    \cr 
     UniTrack \cite{unitrack}          &\xmark & \cmark & \xmark   &-& -& -& -& 58.4& -      \cr   
     STCN$^{-}$ \cite{STCN}         &\xmark & \cmark & \xmark & - & - &- &  \textcolor{blue}{\textbf{82.5}} & 79.3 &   \textcolor{red}{\textbf{85.7}}    \cr 
     SSTVOS\cite{SSTVOS}        &\xmark & \cmark & \xmark  & - &- &- & \textcolor{blue}{\textbf{82.5}} & \textcolor{blue}{\textbf{79.9}} & 85.1   \cr  
     SWEM$^{-}$ \cite{SWEM}     &\xmark & \cmark & \xmark & \textcolor{blue}{\textbf{89.5}} &- &- &81.9 &- &-   \cr
     RTS \cite{RTS}     &\cmark & \cmark & \xmark & - &- &- &80.2 & 77.9 & 82.6  \cr
     \Xhline{\arrayrulewidth}
     OSMN \cite{OSMN}               &\xmark &\xmark & \xmark &73.5 &74.0& 72.9& 54.8& 52.5& 57.1       \cr  
     FAVOS \cite{favos}                &\xmark & \xmark & \xmark &81.0 & 82.4 & 79.5& 58.2& 54.6& 61.8     \cr  
     VideoMatch \cite{videomatch}    &\xmark & \xmark & \xmark   &- &81.0& -& 56.5& -& -     \cr   
     SiamMask \cite{SiamMask}     &\xmark & \xmark & \xmark  & 69.8 & 71.7& 67.8& 56.4& 54.3& 58.5   \cr  
     D3S \cite{D3S}           &\xmark & \xmark & \xmark & 74.0 &75.4& 72.6& 60.8& 57.8& 63.8      \cr  
     Siam R-CNN \cite{D3S}           &\xmark & \xmark & \xmark & - &- & -& 70.6 & 66.1 & 75.0      \cr  
     Unicorn \cite{unicorn}          &\xmark & \xmark & \xmark    &87.4 &86.5& 88.2& 69.2& 65.2& 73.2  \cr  
     OneTracker \cite{hong2024onetracker} &\xmark & \xmark & \xmark & 88.9 & 88.1 & 89.7 & 82.5 & 79.4 & 85.6 \cr
     \textbf{DropSeg} 			   &\xmark & \xmark & \xmark & \textcolor{red}{\textbf{92.1}} & \textcolor{red}{\textbf{90.9}} & \textcolor{red}{\textbf{93.3}} & \textcolor{red}{\textbf{83.0}} & \textcolor{red}{\textbf{80.2}} & \textcolor{red}{\textbf{85.7}}  \cr 
   \Xhline{\arrayrulewidth}  
   \end{tabular}}
   \centering
  \vspace{-0.25cm}
  \label{vos_compare}
\end{table}

 \begin{figure}
\begin{center}
   \includegraphics[width=0.85\linewidth]{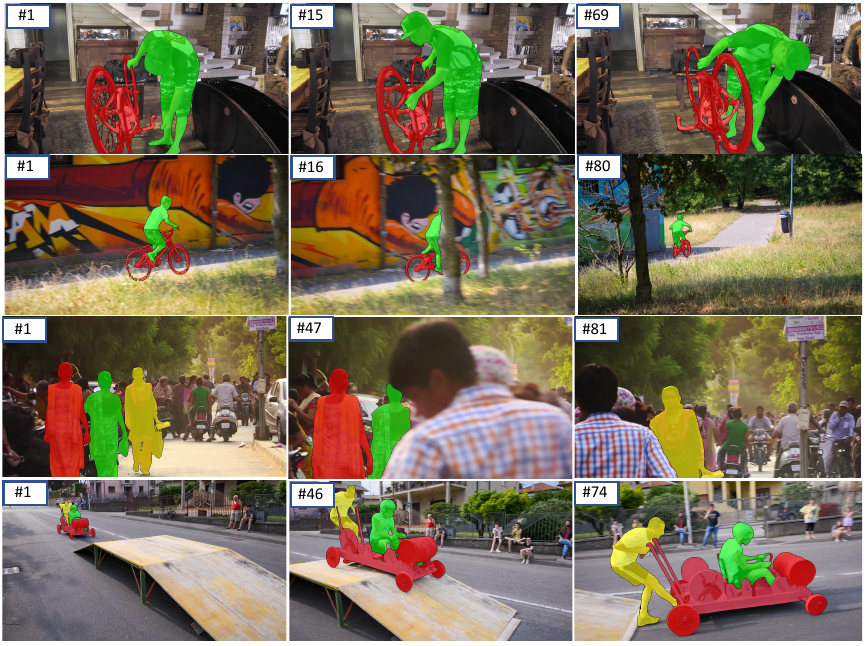} 
\end{center}
\vspace{-0.5cm}
 \caption{Qualitative VOS results of our one-shot approach DropSeg on four sequences in DAVIS-17 \cite{davis17}, which are respectively \emph{bike-packing}, \emph{bmx-trees},  \emph{india} and \emph{soapbox}. The frame number is shown in the \emph{top-left} of each frame, and the ground-truth mask is given in the first frame. 
 }
\label{vos_qualitative}
\end{figure}

\noindent\textbf{Optical flow estimation.} We use our DropMAE pre-trained model as the feature encoder in RAFT \cite{raft}. The extracted features of DropMAE are further upsampled to $2\times$ to match the spatial resolution of RAFT's original features. For training, we first pretrain our DropRAFT on FlyingChairs \cite{flownet}, and then train it on FlyingThings3D \cite{mayer2016large}, following the same training steps and hyper-parameters in \cite{raft}.

\noindent\textbf{Long-term point tracking.} We use our DropMAE pre-trained model as the feature extractor (i.e., Delta-DINO) in \cite{tumanyan2025dino} for online pixel-level correspondence learning, and add one $1\times1$ convolution layer to address the channel dimension mismatch between DropMAE and the original Delta-DINO. We use the same training hyper-parameters and CNN-refiner with DINO-Tracker for fair comparison. For DropMAE, we adopt the LoRA training for parameter-efficient fine-tuning, which is detailed in Table \ref{tap_davis_480}. Following \cite{tumanyan2025dino}, we set lora\_alpha=0.5, lora\_dropout=0.1, rank=8 for LoRA training.


\subsection{Video Object Tracking}
To demonstrate the effectiveness of the proposed DropMAE for VOT, we compare our  DropTrack with state-of-the-art trackers on 7 challenging tracking benchmarks.

\noindent\textbf{GOT-10k.} GOT-10k \cite{got10k} is a challenging dataset that follows the one-shot evaluation protocol, where the trackers are required to be trained on its training split, and the test object classes have no overlap with the objects in the training split. As shown in Table~\ref{overall_results}, our DropTrack achieves state-of-the-art results on this dataset, outperforming OSTrack by 2.2\% and 3.6\% in terms of AO and SR$_{0.5}$. This implies that the temporal correspondence learning in the pre-training is beneficial for the downstream tracking task. Although there exists a domain gap between the pre-training data and the test data (i.e., a large portion of test objects in GOT-10k are animals, vehicles and object parts, whereas K700 only consists of human-centric action videos), the  temporal matching ability learned by DropMAE can still be transferred to the downstream tracking task, improving the tracking performance.

\begin{table}[t]
  \caption{3D tracking results on KITTI-Van w/ limited training samples. Methods equipped with different types of backbones and pre-trained models are included for comparison.} 
  \vspace{-0.3cm}  
  \newcommand{\tabincell}[2]
    \centering
  \footnotesize
    \begin{tabular}{c@{}c@{}cccc}
    \Xhline{\arrayrulewidth}
   \multirow{1}{*}{Method}& Backbone & Pre-Train & Type & Succ. & Prec. \cr
       \Xhline{\arrayrulewidth}  
       P2B \cite{qi2020p2b} &PointNet++ & - & - & 40.8 & 48.4 \cr
       BAT \cite{bat} & PointNet++ & - & - & 52.4 & 67.0 \cr
       DMT \cite{dmt} & PointNet++ & - & - & 53.3 & 65.6 \cr
       M2Track \cite{m2track} &PointNet& - & - & 53.8 & 70.7 \cr
       STNet \cite{stnet} &Transf.& - & - & 58.0 & 70.6 \cr
      MBPTrack \cite{mbptrack} & Transf. & - & - & 61.3 & 72.7 \cr
      SiamDisst \cite{wu2023boosting} & ViT & Recon \cite{recon} & 3D & 62.9 & 73.6 \cr
     SiamDisst \cite{wu2023boosting} & ViT & Point-MAE \cite{pang2022masked} & 3D & \textbf{63.5} & \textbf{75.0} \cr
    \Xhline{\arrayrulewidth}
     SiamDisst \cite{wu2023boosting} & ViT & MAE \cite{mae} & 2D & 60.5 & 69.9  \cr
     SiamDisst \cite{wu2023boosting} & ViT & \textbf{DropMAE} & 2D & \textbf{61.9} & \textbf{74.0} \cr
   \Xhline{\arrayrulewidth}
   \end{tabular}  
   \centering
  \label{3d_track}
\end{table}

\noindent\textbf{LaSOT.} LaSOT consists of 280 long test sequences, and our results are presented in Table \ref{overall_results}.
Our DropTrack sets a new record on this dataset with 71.8\% AUC, 81.8\% P$_{Norm}$ and 78.1\% P, which shows the great potential of our DropTrack in robust long-term visual tracking.

\noindent\textbf{LaSOT$_{ext}$.} LaSOT$_{ext}$ is an extension of LaSOT with more challenging video sequences for testing. Similar to GOT-10k, the test split has a large gap with the training split, and sequences with novel object classes (i.e., not present in ImageNet) are used for evaluation. 
Our DropMAE outperforms the other 
trackers by large margins. Specifically, without a complex memory design,  DropTrack outperforms temporal updating-based ARTrack \cite{artrack} by 0.8\%, 1.9\% and 1.7\% in terms of AUC, P$_{Norm}$ and P metrics. This shows that a tracker with DropMAE pre-training generalizes well to unseen objects in generic visual object tracking.

\noindent\textbf{TNL2K.} TNL2K is a large-scale evaluation dataset that consists of 700 test videos with various challenges, such as significant appearance variation and manually added adversarial samples. As illustrated in Table \ref{overall_results}, our DropMAE significantly outperforms the other trackers on this dataset.

\noindent\textbf{ITB, TrackingNet and OTB100.} In Table \ref{otb_like_compare}, we evaluate our DropTrack on ITB \cite{itb}, OTB100 \cite{OTB100} and TrackingNet \cite{trackingnet}, achieving state-of-the-art performance on each one.
DropTrack is slightly better than OSTrack on ITB and TrackingNet. We believe the main reason is the fully overlapped training and test object classes in these two datasets, which reduces the effect of pre-training. A competitive tracker on these two datasets can be learned even using supervised ImageNet weights, which has been shown in \cite{ostrack}.

{On all 7 VOT datasets, our DropTrack outperforms the baseline OSTrack, which demonstrates that our DropMAE pre-training on videos learns better temporal-matching representations than the MAE model trained on ImageNet, resulting in more a robust tracker that generalizes well to  both unseen and seen objects.}

\noindent\textbf{Visualization.} In Fig. \ref{vot_qualitative}, we show the qualitative tracking results obtained by our DropTrack and the other 3 compared trackers. The selected sequences contain various challenges including significant appearance variation, background cluster, illumination variation and similar objects. Our DropTrack handles these challenges well due to the robust DropMAE pre-trained model.

\begin{table*}[t]
\vspace{-0.3cm}
  \caption{Comparison with previous self-supervised learning approaches on video object segmentation (DAVIS-2017) and human pose propagation (JHMDB) tasks. ``Adaptor'' means that we add the lightweight adaptor on top of the frozen backbone, and only the adaptor is learnable. For all the ViT models, we use a patch size of $16\times16$ with a stride of 8. The best two results are shown in \textcolor{red}{\textbf{red}} and \textcolor{blue}{\textbf{blue}}, respectively.}  
  \vspace{-0.3cm}
  \newcommand{\tabincell}[2]
    \centering
    \begin{tabular}{ccc|ccc|cc}
    \Xhline{\arrayrulewidth}
   \multirow{2}{*}{Method}& \multirow{2}{*}{Backbone}& \multirow{2}{*}{Dataset}&\multicolumn{3}{c|}{DAVIS-2017 \cite{davis17}}&\multicolumn{2}{c}{JHMDB \cite{jhmdb}} \cr
     &  & &  $\mathcal{J}$\&$\mathcal{F}$ & $\mathcal{J}$ & $\mathcal{F}$ & PCK@0.1 & PCK@0.2 \cr
     \hline
     Supervised \cite{resnet} & ResNet-50 & ImageNet & 66.0 & 63.7 & 68.4 &  59.2 & 78.3 \cr
     TimeCycle \cite{timecycle} & ResNet-50 & VLOG & 40.7 & 41.9 & 28.9 & 57.7 & 78.5 \cr
     UVC \cite{uvc} & ResNet-18 & K400 & 57.8 & 56.3 & 59.2 &  58.6 & 79.6 \cr
     SimSiam \cite{chen2021exploring} &  ResNet-50 & ImageNet & 66.3 & 64.5 & 68.2 & 58.4 & 77.5 \cr
     MoCo \cite{moco} & ResNet-50 & ImageNet & 65.4 & 63.2 & 67.6 &   \textcolor{blue}{\textbf{60.4}} & 79.3 \cr
     VINCE \cite{gordon2020watching} &  ResNet-50 & K400 & 65.6 & 63.4 & 67.8 & 58.2 & 76.3 \cr
     RegionTracker \cite{purushwalkam2020demystifying} & ResNet-50 & TrackingNet & 63.4 & 61.5 & 65.4 & 57.5 & 74.6 \cr
     CRW \cite{crw} & ResNet-18 & K400 & 67.6 & 64.8 & 70.2 & 59.3 & 80.3 \cr
     DUL \cite{DUL} & ResNet-18 & YT-VOS & \textcolor{blue}{\textbf{69.3}} & \textcolor{red}{\textbf{67.1}} & 71.6 & 56.4 & 79.1 \cr
     DUL \cite{DUL} & ResNet-18 & K400 & 68.7 & \textcolor{blue}{\textbf{66.7}} & 70.7 & 58.2 & 80.5 \cr
     VFS \cite{vfs} & ResNet-50 & K400 & 68.9 & 66.5 & 71.3 &  \textcolor{red}{\textbf{60.9}} &  \textcolor{blue}{\textbf{80.7}} \cr
     \hline
     MAE \cite{mae} & ViT-B/16 & ImageNet & 	59.1 & 57.1 & 61.2 & - & - \cr
     OMNIMAE \cite{girdhar2023omnimae} & ViT-B/16& SSv2+ImageNet & 36.2 & 34.7 &37.6 & - & - \cr
     MME \cite{MME} & ViT-B/16 & ImageNet & 59.2 & 57.1 & 61.2 & - & - \cr
     VideoMAE \cite{videomae} &  ViT-B/16 & K400 & 43.4 &  41.9  & 44.9 & - & - \cr
     DropMAE & ViT-B/16 & K400 & 60.3 & 58.3 &62.3 & - & - \cr
    \hline
    VideoMAE \cite{videomae} + Adaptor	 & ViT-B/16 & YT-VOS & 57.7 & 54.9 & 60.6 & - & - \cr
    DropMAE + Adaptor & ViT-B/16 & YT-VOS & 	\textcolor{red}{\textbf{69.4}} & 66.3 & \textcolor{red}{\textbf{72.5}} & 57.3	& 80.2 \cr
    DropMAE + Adaptor & ViT-B/16 & K400 & 	68.7 & 65.5 &\textcolor{blue}{\textbf{71.9}} & 	57.8	& \textcolor{red}{\textbf{80.8}} \cr 
       \Xhline{\arrayrulewidth}  
   \Xhline{\arrayrulewidth}  
   \end{tabular}  
   \centering
  \label{corres_leanring}
\end{table*}

\begin{table}[t]
\vspace{-0.5cm}
  \caption{Training efficiency comparison with previous self-supervised approaches. With the frozen DropMAE model, fine-tuning only an adaptor with fewer learnable parameters (5.4M) achieves comparable performance to the DUL baseline, while speeding up training by 4.6× to 16.6× on YT-VOS and K400 datasets, respectively. The best efficiency is in \textbf{bold}. The training time is measured on a single RTX 3090 GPU.}  
  \vspace{-0.3cm}
  \newcommand{\tabincell}[2]
    \centering
  \footnotesize
    \begin{tabular}{cccc}
    \Xhline{\arrayrulewidth}
   \multirow{1}{*}{Method}& \multirow{1}{*}{Learn. Model}& \multirow{1}{*}{Training Time}&\multirow{1}{*}{Dataset} \cr
       \Xhline{\arrayrulewidth}  
      CRW \cite{crw} & ResNet-18 (11.2M) & 168 Hours & K400  \cr
      DUL \cite{DUL} & ResNet-18 (11.5M) & 182.9 Hours & K400 \cr
      DUL \cite{DUL} & ResNet-18 (11.5M) & 16 Hours & YT-VOS \cr
      DropMAE + Adaptor &  Adaptor (5.4M) &\textbf{11} Hours & K400 \cr 
      DropMAE + Adaptor &  Adaptor (5.4M) & \textbf{3.5} Hours & YT-VOS  \cr
   \Xhline{\arrayrulewidth}
   \end{tabular}  
   \centering
   \vspace{-0.3cm}
  \label{corres_compex}
\end{table}

\begin{table}[t]
\vspace{-0.5cm}
  \caption{Comparison with previous self-supervised approaches on OTB100 \cite{OTB100}. The best results are shown in \textbf{bold}.}
  \vspace{-0.3cm}  
  \newcommand{\tabincell}[2]
    \centering
  \footnotesize
    \begin{tabular}{cccc}
    \Xhline{\arrayrulewidth}
   \multirow{1}{*}{Method}& \multirow{1}{*}{Backbone}& \multirow{1}{*}{Dataset}&\multirow{1}{*}{AUC} \cr
       \Xhline{\arrayrulewidth}  
      Supervised \cite{resnet} & ResNet-50 & ImageNet & 45.5\cr
      SimSiam \cite{chen2021exploring} & ResNet-50 & ImageNet & 43.2 \cr
      MoCo \cite{moco} &ResNet-50 & ImageNet & 46.5\cr	
      VINCE \cite{gordon2020watching} &ResNet-50 & K400 & 47.6 \cr	
      RegionTracker \cite{purushwalkam2020demystifying} & ResNet-50 & TrackingNet &  43.4 \cr	
      SeCo \cite{yao2021seco} & ResNet-50 & K400 &  51.8 \cr	
      VFS \cite{vfs} & ResNet-50 & K400 & 43.4 \cr
      VFS \cite{vfs} &	ResNet-50 & K400+GOT-10k & 52.5 \cr
      VideoMAE \cite{videomae} + Adaptor & ViT-B/16 &  K400 & {43.5} \cr	
      DropMAE + Adaptor & ViT-B/16 &  K400 & \textbf{53.2} \cr
   \Xhline{\arrayrulewidth}
   \end{tabular}  
   \centering
  \label{corres_otb}
\end{table}

\begin{table}[t]
\vspace{-0.4cm}
  \caption{Optical flow estimation results on Sintel (train). We train our DropRAFT on FlyingChairs \cite{flownet} and FlyingThings3D \cite{mayer2016large}, and test it on Sintel (train) for generalization performance evaluation.}
  \vspace{-0.3cm}  
  \newcommand{\tabincell}[2]
    \centering
  \footnotesize
    \begin{tabular}{cc|cc}
    \Xhline{\arrayrulewidth}
   \multirow{1}{*}{Method}& \multirow{1}{*}{Source}& Clean $\downarrow$ & Final $\downarrow$ \cr
       \Xhline{\arrayrulewidth}  
      HD3 \cite{yin2019hierarchical} & CVPR19 &  3.84 & 8.77 \cr
      FlowNet2 \cite{flownet} &CVPR17 &  2.02 & 3.54 \cr
     PWC-Net \cite{sun2018pwc} & TPAMI19 & 3.45 & 4.60 \cr
     GMA \cite{gmflow} &ICCV21 &  1.30 & 2.74 \cr
     GMFlow \cite{gmflow} &ICCV21 & 1.08 & 2.48 \cr
     SKFlow \cite{sun2022skflow} &NeurIPS22 & 1.22 & 2.46 \cr
     DIP \cite{zheng2022dip} &CVPR22 & 1.30 & 2.82 \cr
     CRAFT \cite{sui2022craft} & CVPR22 & 1.27 & 2.79 \cr
     RAFT-it \cite{sun2022disentangling} & ECCV22 & 1.74 & 2.41 \cr
     GMFlowNet \cite{zhao2022global} &CVPR22 &  1.14 & 2.71 \cr
     FlowFormer++ \cite{shi2023flowformer++} & CVPR23 & 0.94 & 2.33 \cr
     TransFlow \cite{lu2023transflow} & CVPR23 &  0.93 & 2.33 \cr
     EMD-L \cite{deng2023explicit} & ICCV23 & \textbf{0.88} & 2.55 \cr
     RPKNet \cite{morimitsu2024recurrent} & AAAI24 & 1.12 & 2.45 \cr
     SEA-RAFT (M) \cite{searaft} & ECCV24 &  1.21 & 4.04 \cr
     SEA-RAFT (L) \cite{searaft} & ECCV24 & 1.19 & 4.11 \cr
    \hline
    RAFT \cite{raft} & ECCV20 &  1.43 & 2.71 \cr
    ViT-RAFT & Random & 13.56 & - \cr
    ViT-RAFT & VideoMAE \cite{videomae} & 13.48 & - \cr
    ViT-RAFT & MAE \cite{mae} & 1.51 & 2.72 \cr
    \textbf{DropRAFT} & Ours & 1.06  & \textbf{2.25}  \cr
   \Xhline{\arrayrulewidth}  
   \end{tabular}  
   \centering
  \vspace{-0.35cm}
  \label{comp_optical_flow}
\end{table}

\begin{table}[t]
  \caption{Long-term point tracking performance on TAP-Vid DAVIS-480 \cite{doersch2022tap}.  Our test-time self-supervised trackers perform favourably against supervised methods \cite{karaev2025cotracker,doersch2023tapir} trained w/ large-scale annotated data, while outperforming SOTA  test-time DINO-Tracker by using fewer learnable parameters. ``Param.'' means learnable backbone parameters during the test-time training. Modes `S' and `TT' indicate supervised and test-time training, respectively.}
  \vspace{-0.3cm}
  \newcommand{\tabincell}[2]
    \centering
  \footnotesize
    \begin{tabular}{@{}c@{}c@{\hspace{0.1cm}}c@{}c|ccc@{}}
    \Xhline{\arrayrulewidth}
    \multirow{1}{*}{Method}& Mode &  
  \multirow{1}{*}{Param.}& \multirow{1}{*}{Backbone}&\multirow{1}{*}{$\delta^{x}_{avg}$} &\multirow{1}{*}{OA} &\multirow{1}{*}{AJ}
\cr
       \Xhline{\arrayrulewidth}  
     RAFT \cite{raft} && - & CNN & 66.7 & - & -  \cr
     DINOv2 \cite{oquab2023dinov2}            && -  & ViT-L-14/7 & 66.7 & - & -      \cr  
     \hline
     TAP-Net \cite{doersch2022tap} & S & -  & CNN & 66.4 & 79.0 & 46.0 \cr
     PIPs++ \cite{zheng2023pointodyssey} & S & -  & CNN & 73.6 & - & - \cr
     TAPIR \cite{doersch2023tapir} & S & -  & CNN & 77.3 & 89.5 & 65.7 \cr
     Co-Tracker \cite{karaev2025cotracker} & S & -  & Transf.& 79.4 & 89.5 & 65.6 \cr
     \hline
     Omnimotion \cite{wang2023tracking} & TT & -   & MLP & 74.1 & 84.5 & 58.4 \cr
     DINO-Tracker \cite{tumanyan2025dino} & TT & 7.59M & CNN  &  \textcolor{red}{\textbf{80.4}}  & 88.1 & 64.6 \cr
     DINOv2$^{LoRA-2L}$ \cite{tumanyan2025dino} & TT  & 0.07M & ViT-L-14/7  & 73.2 & 84.8 & 58.0 \cr
     \textbf{DropDINO}$^{LoRA-2L}$ & TT & 0.05M   & ViT-B-16/8 &79.0 & 89.1 & 64.9 \cr
     \textbf{DropDINO}$^{LoRA-4L}$ & TT & 0.10M &  ViT-B-16/8  & 79.5 & \textcolor{blue}{\textbf{89.6}}  & \textcolor{blue}{\textbf{65.4}}\cr
     \textbf{DropDNO}$^{LoRA-6L}$ & TT & 0.15M &  ViT-B-16/8   & \textcolor{blue}{\textbf{79.7}} & \textcolor{red}{\textbf{89.8}}& \textcolor{red}{\textbf{65.7}} \cr
   \Xhline{\arrayrulewidth}  
   \end{tabular}  
   \centering
  \label{tap_davis_480}
\end{table}

\begin{figure}
\vspace{-0.2cm}
\begin{center}
   \includegraphics[width=1.0\linewidth]{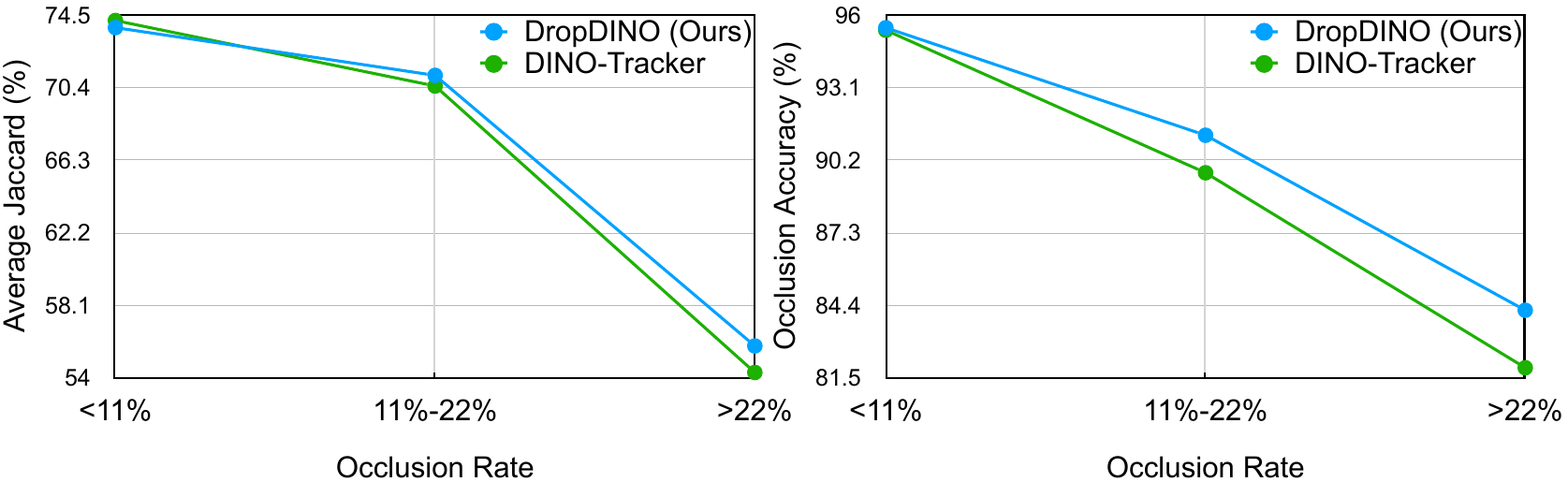} 
\end{center}
\vspace{-0.55cm}
 \caption{\jimmyy{Tracking performance on TAP-Vid DAVIS-480 \cite{doersch2022tap} with varied occlusion rate. We divide the test videos from TAP-Vid DAVIS-480 into three groups according to the average occlusion rate of each video, which is estimated using ground-truth visibility annotations. Average Jaccard (AJ) and Occlusion Accuracy (OA) are reported. Our DropDINO exhibits less performance degradation as the occlusion rate increases (i.e., $>22\%$), demonstrating its potential in long-term point tracking with severe occlusion.}
 }
\label{occlusion_plot}
\vspace{-0.2cm}
\end{figure}

\subsection{Video Object Segmentation}
In Table \ref{vos_compare}, we compare our DropSeg with existing VOS approaches on the DAVIS-16/17 \cite{davis16,davis17}.

\noindent\textbf{DAVIS-16.} DAVIS-16 is composed of 20 manually annotated test sequences. As shown in Table \ref{vos_compare}, our one-shot DropSeg approach, without using any online learning and complicated memory mechanisms, achieves the best  $\mathcal{J}\&\mathcal{F}$ score of 92.1\%, which significantly outperforms the other compared one-shot approaches and is even better than the approaches with complicated pipelines (i.e., OL, M and S). This implies that the pixel-wise correspondence learned during the pre-training is effective for capturing long-range dependencies between various frames in VOS.

\noindent\textbf{DAVIS-17.} DAVIS-17 is an extension of DAVIS-16, comprising more challenging videos and supports multi-object segmentation. In Table \ref{vos_compare}, our DropSeg  achieves competitive results of 83.0\% $\mathcal{J}\&\mathcal{F}$, 80.2\% $\mathcal{J}$ and 85.7\% $\mathcal{F}$, which shows its superiority in handling more challenging videos.

\noindent\textbf{Visualization.} The qualitative visualization of our DropSeg is shown in Fig.~\ref{vos_qualitative}. Even without using online fine-tuning or complicated memory mechanisms, our DropSeg can still provide accurate segmentation results in the following frames by only using the mask annotation in the first frame, which is mainly due to the favorable temporal matching ability learned in DropMAE. 


\vspace{-0.3cm}
\subsection{3D Point Cloud Tracking}
In Table \ref{3d_track}, we evaluate our DropMAE pre-trained model in 3D point cloud tracking. Our DropMAE variant outperforms the other transformer-based 3D trackers (e.g., MBPTrack \cite{mbptrack} and STNet \cite{stnet}) in terms of both success and precision metrics. Notably, the performance achieved by DropMAE is even comparable to 3D pre-training approaches (i.e., Point-MAE \cite{pang2022masked} and Recon \cite{recon}), which demonstrates that the temporal matching ability learned from 2D videos can be well transferred to 3D tracking.

\vspace{-0.3cm}
\subsection{Self-Supervised Correspondence Learning}
In Table \ref{corres_leanring}, we treat DropMAE as the frozen feature extractor, and evaluate its unsupervised VOS and pose propagation performance on DAVIS-2017 \cite{davis17} and JHMDB \cite{jhmdb}. 
DropMAE achieves better performance than the other generative models (e.g., MAE, OMNIMAE and MME) on both two tasks, and significantly outperforms the video-based generative approach VideoMAE. This is mainly because VideoMAE adopts 3D CNN for cube extraction along the temporal dimension, which does not learn effective temporal correspondence and is more suitable for high-level video action recognition task. 

In addition, we add one lightweight adaptor on top of the frozen DropMAE feature extractor, and train the adaptor via \cite{DUL}. Our new variant achieves comparable unsupervised tracking performance to the  DUL baseline \cite{DUL}, while respectively running $4.6\times$ and $16.6\times$ faster training on YT-VOS and and K400, as illustrated in Table \ref{corres_compex}. Notably, we evaluate the representations learned by our DropMAE + Adaptor on the traditional object-level unsupervised tracking task (see Table \ref{corres_otb}) and our DropMAE achieves the leading performance.
 
 \begin{figure}
\begin{center}
   \includegraphics[width=0.9\linewidth]{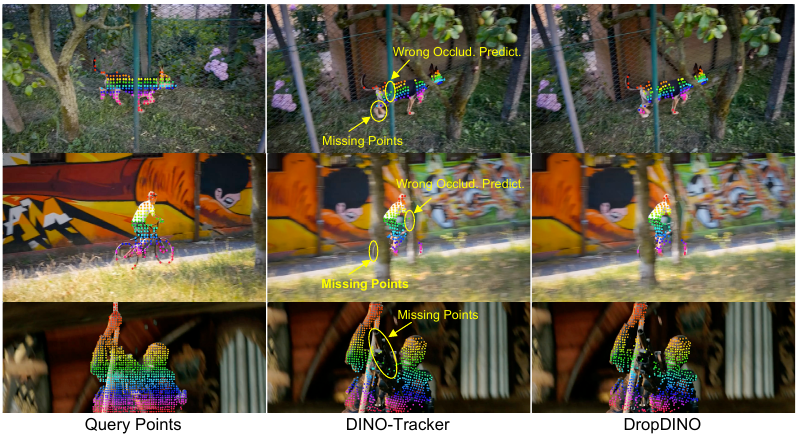} 
\end{center}
\vspace{-0.65cm}
 \caption{Long-term point tracking results obtained by DINO-Tracker \cite{tumanyan2025dino} and our {DropDNO}$^{LoRA-6L}$. Our approach can better handle target occlusion and achieves more robust dense point tracking.
 }
\label{point_qualitative}
\vspace{-0.3cm}
\end{figure}

\vspace{-0.3cm}
\subsection{Optical Flow Estimation}
Following the standard evaluation \cite{raft}, we train our DropRAFT on the FlyingChairs \cite{flownet} and FlyingThings \cite{mayer2016large} synthetic datasets, and test the zero-shot generalization performance on the training set of Sintel \cite{sintel}. As shown in Table \ref{comp_optical_flow}, our DropRAFT significantly outperforms the baseline RAFT by large margins on both Clean and Final splits, which demonstrates the effectiveness of our DropMAE backbone in optical flow estimation. Compared to MAE-RAFT using the same ViT-B/16 backbone, DropRAFT obtains lower average endpoint error, which indicates that  DropMAE is a better temporal learner than MAE. FlowFormer++ applies mask autoencoding in optical flow estimation. Although it is specifically designed for optical flow, our general DropMAE still outperforms it on the Final split, showing the generalization of the DropMAE's representation to this task. \jimmyy{For ViT-RAFT using Random and VideoMAE initialization, we observe that they suffer from severe training collapse. Given the success of MAE and our DropRAFT, pre-trained weights are essential for ViT-based optical flow estimation.}




\vspace{-0.3cm}
\subsection{Long-term Point Tracking}
As illustrated in \S\ref{sec:point_track}, we use our DropMAE as the feature extractor in the DINO-Tracker framework. For parameter-efficient training, we adopt the LoRA training, which specifically optimizes the last $N$ layers (i.e., denoted as $NL$ in Table \ref{tap_davis_480}) of the backbone. From Table \ref{tap_davis_480}, we can observe that our DropDINO$^{LoRA-2L}$, which only optimizes the last two layers of our DropMAE w/ fewer learnable parameters (0.07M), outperforms the SOTA test-time optimization-based DINO-Tracker \cite{tumanyan2025dino} in terms of both Occlusion Accuracy (OA) and Average Jaccard (AJ) metrics. 
\jimmyy{For the point accuracy ($\delta^{x}_{avg}$) metric, our DropDINO achieves comparable performance to DINO-Tracker even using a coarse ViT patch embedding layer with a large patch size of $16\times16$, while DINO-Tracker uses a more find-grained CNN kernel size of $7\times7$.} 
Adding more layers for LoRA training in DropDINO leads to consistent improvements in terms of all metrics, with a relatively small increase in the number of parameters.  In addition, \cite{tumanyan2025dino} uses DINOv2 as the backbone for LoRA training (denoted as DINOv2$^{LoRA-2L}$$^{\dag}$) and suffers from performance degradation, which is mainly due to the lack of temporal prior in DINOv2. Compared with approaches that use large-scale datasets for supervised training (e.g., Co-tracker \cite{karaev2025cotracker} and TAPIR \cite{doersch2023tapir}), our {DropDINO}$^{LoRA-6L}$ achieves comparable performance but using fewer learnable parameters (\jimmyy{see Fig. \ref{occlusion_plot}}), which shows its potential to be served as a general backbone for long-term dense point tracking.

\noindent\textbf{Visualization.} The qualitative point tracking results of our DropDINO and DINO-Tracker are shown in Fig. \ref{point_qualitative}. Our approach  more effectively handles target occlusion (\jimmyy{see Fig. \ref{occlusion_plot}}) and accurately tracks dense points, which is mainly due to the effective temporal learner (i.e., DropMAE) in our DropDINO.

\begin{figure}[t]
\begin{center}
   \includegraphics[width=0.88\linewidth]{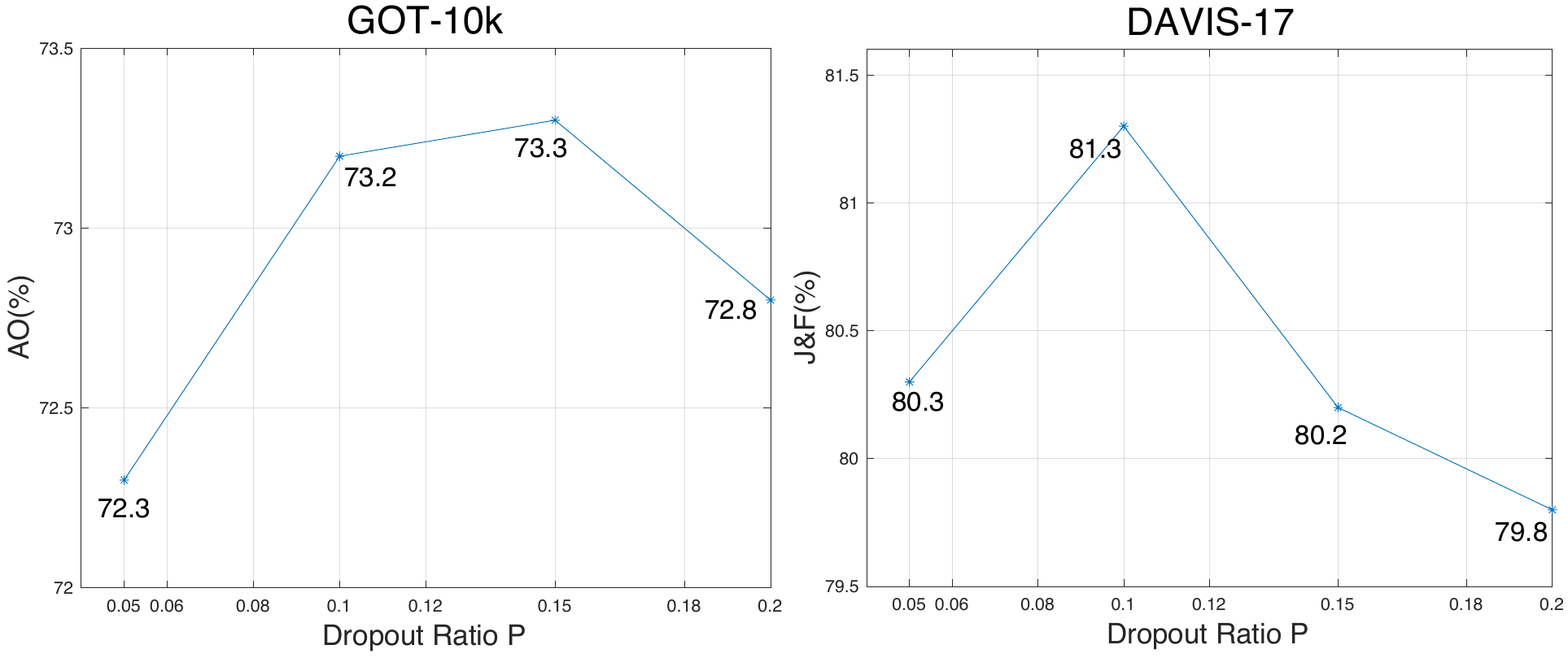} 
\end{center}
\vspace{-0.65cm}
 \caption{Ablation study of the dropout ratio $P$ in DropMAE on the GOT-10k (VOT) and DAVIS-17 (VOS) datasets.
 }
\label{abl_p}
\vspace{-0.35cm}
\end{figure}

\vspace{-0.35cm}
\section{Ablation Studies}
\label{sec:abl}
In this section, we conduct  ablation studies to provide more detailed analysis of our method.
We use DropMAE with 400-epoch pre-training for the ablation study.

\begin{table}[t]
  \newcommand{\tabincell}[2]
    \centering
    \footnotesize
      \caption{The downstream VOT and VOS performance on GOT-10k 
  and DAVIS-17
  obtained by using our DropMAE pre-trained on various video datasets. VOS uses 800 epochs pre-training.}
  \vspace{-0.35cm}
    \begin{tabular}{ccc|c@{\hspace{0.1cm}}c@{\hspace{0.1cm}}c|c}
    \Xhline{\arrayrulewidth}
    \multirow{2}{*}{Datasets}&  
  No.& No. &\multicolumn{3}{c|}{VOT} &\multicolumn{1}{c}{VOS} \cr
  & Videos & Actions& 
  AO & SR$_{0.5}$ & SR$_{0.75}$ & $\mathcal{J}\&\mathcal{F}$ \cr
     \Xhline{\arrayrulewidth}  
     K400 \cite{k400}             &240,000   &400 &73.2&83.9&67.5 & 82.7     \cr  
     K600 \cite{k600}             &390,000   &600 &74.5&85.5&69.5  & 82.8     \cr  
     K700 \cite{k700}             &526,768   &700 &\textbf{75.6}&\textbf{86.2}&\textbf{71.4} & \textbf{83.0}      \cr  
     MiT \cite{moments}             &802,244   &339 &75.1&85.5&70.6  & 82.8     \cr  
     WebVid \cite{webvid}             &240,000   &- &72.8&83.4&67.3   & 81.5   \cr   
     WebVid \cite{webvid}              &960,000   &- &73.4&85.0&69.5   & 82.9   \cr  
   \Xhline{\arrayrulewidth}  
   \end{tabular}  
   \centering
  \vspace{-0.3cm}
  \label{video_source}
\end{table}

\begin{table}[t]
  \newcommand{\tabincell}[2]
    \centering
    \footnotesize
      \caption{The tracking performance of AO/SR$_{0.75}$ on GOT10-k reported by variants with different settings.}
      \vspace{-0.35cm}
    \begin{tabular}{ccccc}
    \Xhline{\arrayrulewidth}
    \multirow{1}{*}{Settings}&  
\multicolumn{1}{c}{GOT-10k}\cr
     \Xhline{\arrayrulewidth}  
     {DropTrack-K400-400E} & 73.2/67.5 \cr
     w/ ASAD in Encoder  & 73.1/68.1 \cr  
     w/ domain specific data & 73.4/68.8\cr 
     w/o frame identity embed & 72.9/67.4 \cr  
   \Xhline{\arrayrulewidth}  
   \end{tabular}  
   \centering
  \label{abl_factors}
\end{table}

\begin{table}[t]
\vspace{-0.3cm}
  \newcommand{\tabincell}[2]
    \centering
      \caption{The effect of maximum sampling frame gap on the downstream tracking task.}  
      \vspace{-0.35cm}
  \footnotesize
    \begin{tabular}{c|ccc}
    \Xhline{\arrayrulewidth}
    \multirow{2}{*}{Maximum Sampling Frame Gap}&  \multicolumn{3}{c}{GOT-10k} \cr
   & AUC & SR$_{0.5}$ & SR$_{0.75}$ \cr

       \Xhline{\arrayrulewidth}  
     1             &72.2 &  82.7 & 65.7  \cr  
    10            &72.8 & 83.4 & 67.2     \cr  
     50           &\textbf{73.2} & \textbf{83.9} & \textbf{67.5}    \cr  
   \Xhline{\arrayrulewidth}  
   \end{tabular}  
   \centering
  \label{samp_gap}
  \vspace{-0.3cm}
\end{table}

\begin{table}[t]
  \newcommand{\tabincell}[2]
    \centering
      \caption{The comparison between DropMAE,  MAE-K400-static and RandDrop-MAE on GOT-10k \cite{got10k}.}  
      \vspace{-0.35cm}
  \footnotesize
    \begin{tabular}{c|ccc}
    \Xhline{\arrayrulewidth}
    \multirow{2}{*}{Methods}&  \multicolumn{3}{c}{GOT-10k} \cr
   & AUC & SR$_{0.5}$ & SR$_{0.75}$ \cr
       \Xhline{\arrayrulewidth}  
     DropMAE              &\textbf{73.2} & \textbf{83.9} & \textbf{67.5}  \cr  
     MAE-K400-static           &70.4 &	80.7 &65.6 \cr  
     RandDrop-MAE           &71.7 &82.4 &66.2 \cr
   \Xhline{\arrayrulewidth}  
   \end{tabular}  
   \centering
  \label{k400_static}
\end{table}

\CUT{
\begin{table}[t]
  \newcommand{\tabincell}[2]
    \centering
      \caption{The comparison between DropMAE and  RandDrop-MAE on GOT-10k \cite{got10k}.}  
      \vspace{-0.3cm}
  \footnotesize
    \begin{tabular}{c|ccc}
    \Xhline{\arrayrulewidth}
    \multirow{2}{*}{Methods}&  \multicolumn{3}{c}{GOT-10k} \cr
   & AUC & SR$_{0.5}$ & SR$_{0.75}$ \cr
       \Xhline{\arrayrulewidth}  
     DropMAE              &\textbf{73.2} & \textbf{83.9} & \textbf{67.5}  \cr  
     RandDrop-MAE           &71.7 &82.4 &66.2
 \cr  
   \Xhline{\arrayrulewidth}  
   \end{tabular}  
   \centering
  \label{rand_drop}
  \vspace{-0.3cm}
\end{table}
}

\begin{table}[t]
\vspace{-0.4cm}
  \newcommand{\tabincell}[2]
    \centering
      \caption{The ablation study on the usage of the pre-trained MAE model to initialize DropMAE pre-training.}  
      \vspace{-0.3cm}
  \footnotesize
    \begin{tabular}{c|ccc|c}
    \Xhline{\arrayrulewidth}
    \multirow{2}{*}{Pre-trained MAE}&  \multicolumn{3}{c|}{GOT-10k} &  \multicolumn{1}{c}{Davis-17} \cr
   & AUC & SR$_{0.5}$ & SR$_{0.75}$ & $\mathcal{J}\&\mathcal{F}$ \cr
       \Xhline{\arrayrulewidth}  
     w/o              &73.2 & {83.9} & {67.5} & 81.3 \cr  
     w/           &\textbf{75.2} & \textbf{85.4} & \textbf{71.5} & \textbf{82.6}
 \cr  
   \Xhline{\arrayrulewidth}  
   \end{tabular}  
   \centering
 \vspace{-0.55cm}
  \label{pretrained_mae}
\end{table}

 \begin{figure}[t]
\begin{center}
   \includegraphics[width=0.9\linewidth]{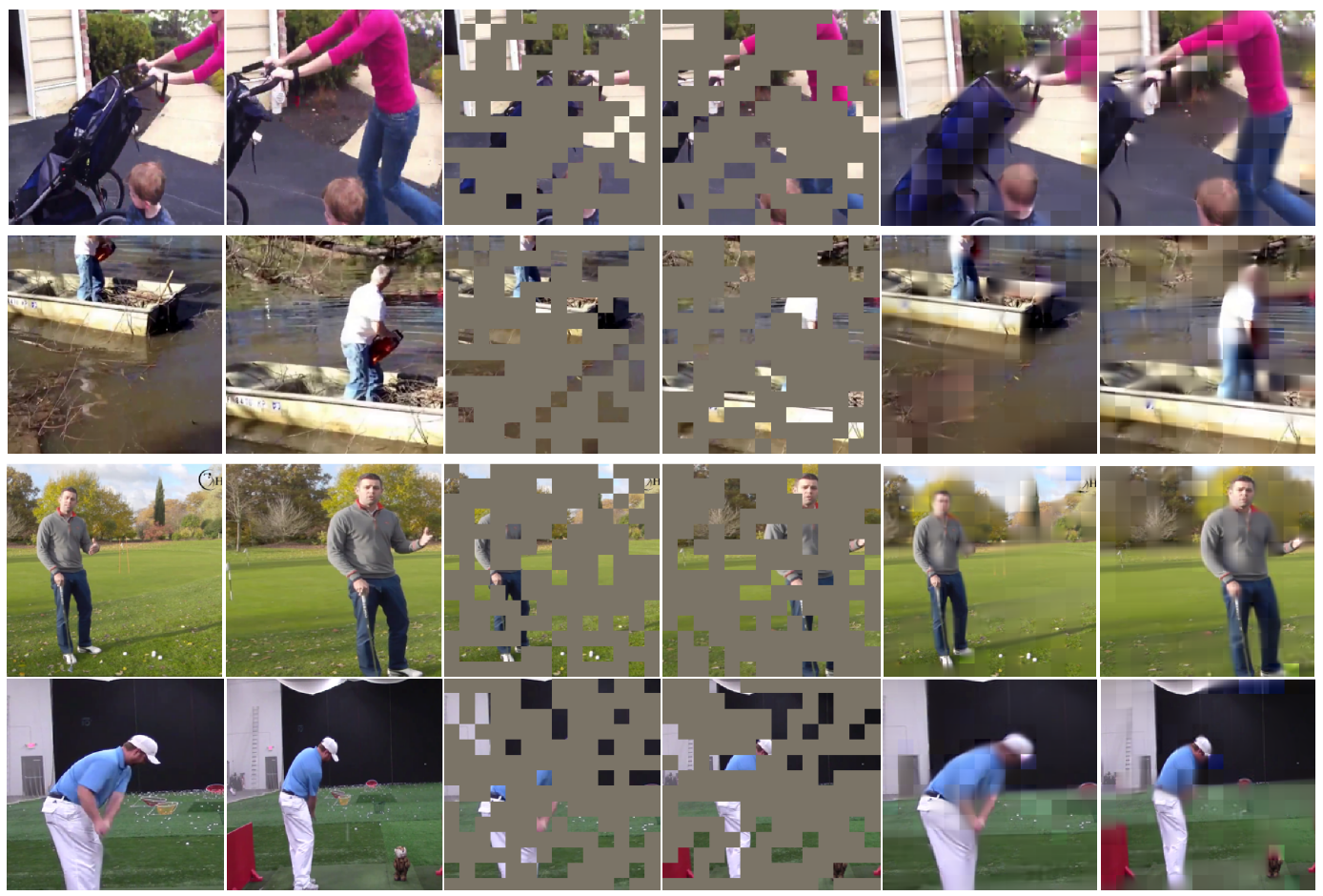} 
\end{center}
\vspace{-0.55cm}
 \caption{Video frame reconstruction results of DropMAE on K400 validation set. We show the original input frame pairs, masked frame pairs (i.e., with 75\% mask ratio) and reconstruction results, sequentially.}
\label{resonstruction}
\vspace{-0.3cm}
\end{figure}

\noindent\textbf{The effect of dropout ratio $P$}. We study the effect of dropout rate $P$ in Fig. \ref{abl_p}. 
A relatively small dropout ratio of $P=0.1$ works well on both VOT and VOS tasks. Meanwhile, dropping too many spatial cues (e.g., $P$=0.2) degrades the downstream tasks, which is mainly because the spatial cues are also useful for accurate localization and segmentation. $P=0.1$ is the optimal setting, and thus we adopt it in the following experiments.

\noindent\textbf{Pre-training video sources.} Since we are the first to explore masked autoencoder pre-training for temporal matching tasks, it is not clear which video dataset is the optimal choice for pre-training. Here, we use five popular video datasets for pre-training,  including K400 \cite{k400}, K600 \cite{k600}, K700 \cite{k700}, MiT \cite{moments} and WebVid-2M \cite{webvid}. For WebVid-2M, we randomly sample 240k and 960k videos for fair comparison and faster validation. 
The downstream tracking results are reported in Table \ref{video_source}. Performance is not favorable even using 960,000 videos in WebVid for pre-training. This indicates that WebViD is not a good choice for tracking pre-training, which is mainly because it is a video caption dataset that focuses on scene diversity and lacks  rich object motion. Using the K400/600/700 or MiT, tracking benefits from pre-training with from rich action classes (i.e., 700 action classes of K700), from which the model can learn stronger temporal matching ability.

\noindent\textbf{Applying ASAD to the encoder.} We test applying ASAD\CUT{also} to all layers {in both the encoder and decoder of the masked autoencoder.}\CUT{ the encoder model.} 
As shown in Table~\ref{abl_factors}, this variant improves over the original baseline  (0.6\% in SR$_{0.75}$). Considering its additional cost and limited  improvement, we only apply ASAD to the decoder. 

\noindent\textbf{Domain specific data.} We also add  tracking training data (without using box annotations), including TrackingNet, LaSOT, and GOT-10k, into K400 for pre-training.
The downstream tracking performance by using the larger pre-training set is 73.4/68.8, which is better than the baseline. It shows that the domain-specific data is helpful to bridge the domain gap, which can be considered as future work to extend Kinetic datasets with more tracking videos.

\noindent\textbf{Frame identity embedding.} During pre-training, 
the frame identity embedding is used to identify masked patches in the same 2D location of the two frames. From Table \ref{abl_factors}, we can find that downstream fine-tuning without the frame identity embedding performs worse than with it, since not using it is inconsistent with the pre-training stage.

\noindent\textbf{Effect of maximum sampling frame gap.} During the pre-training, we randomly sample two frames of a training video with a predefined maximal sampling frame gap $g$. Here, we study its effect on the downstream VOT task. As shown in Table \ref{samp_gap}, the VOT task benefits more from the large sampling frame gap, i.e., 50. This is because the stronger temporal matching ability can be learned by using the relatively large sampling frame gap. Since the limited performance improvements from  $g=10$ to $g=50$, we directly use $g=50$ for all the pre-training experiments.

\noindent\textbf{Learning static frame representation from K400.} To demonstrate the temporal correspondence learning in   pre-training is the key to the success for downstream tracking tasks, we treat K400 \cite{k400} as a static image dataset and perform the original MAE pre-training on it, which we denote as MAE-K400-static. Specifically, in each training iteration, one frame image is randomly sampled of a video for masked autoencoding pre-training. To make a fair comparison with our DropMAE, we double the video number in this baseline such that the total sampled frame number in one epoch training is the same as DropMAE. The comparison between MAE-K400-static and DropMAE is shown in Table \ref{k400_static}. 
Without temporal correspondence learning,  MAE-K400-static is significantly worse than our DropMAE, which further demonstrates the effectiveness of the temporal correspondence learning in the DropMAE pre-training. 

\noindent\textbf{Random dropout.} The vanilla ViT \cite{ViT} implements dropout \cite{dropout} in each multi-head self-attention layer. To see whether this random dropout works in our masked autoencoding pre-training setting, we build a baseline called RandDrop-MAE, which adopts the random dropout in each self-attention layer of the decoder during the pre-training. Different from our adaptive dropout strategy (i.e., ASAD), RandDrop-MAE randomly drops between-frame or within-frame attentions. For a fair comparison, we use the same dropout ratio (i.e., 0.1) for RandDrop-MAE. As shown in Table \ref{k400_static}, RandDrop-MAE degrades the performance compared with our DropMAE. We believe this is because the random dropout may excessively drop some attention elements that are essential for reconstruction and thus degrade the learning.

\noindent\textbf{Pre-trained MAE.}  The downstream VOT and VOS tasks consist of large amounts of objects with diverse classes for evaluation. Considering that K400 is composed of human-action videos, there still exists domain gap between the pre-training and fine-tuning stages. In order to alleviate this gap, we use the original MAE trained on ImageNet as the pre-training weights of our DropMAE, and then we further pre-train our DropMAE on K400 for temporal correspondence learning. From Table \ref{pretrained_mae}, our DropMAE benefits from the pre-trained MAE on both VOT and VOS tasks, which is mainly because the diverse object classes learned in MAE are beneficial for generic object tracking and segmentation. This also shows the potential that the better downstream performance can be achieved by using the pre-trained MAE and larger video data sources (e.g., K700 \cite{k700}).

\noindent\textbf{Frame Reconstruction Visualization.} We show the video frame reconstruction results obtained by our DropMAE in Fig.~\ref{resonstruction}. 
Although less spatial cues are leveraged in the reconstruction, our DropMAE still achieves favorable reconstruction results by exploring temporal cues or between-frame patches, demonstrating the effectiveness of the proposed approach.

\vspace{-0.3cm}
\section{Conclusion}\label{sec:conclusion}
This paper investigated masked autoencoding pre-training for various temporal matching-based downstream tasks, including object-level tracking (i.e., VOT and VOS), pixel-level tracking (i.e., optical flow estimation and long-term point tracking), 3D point cloud tracking, and self-supervised visual correspondence learning. Specifically, we propose an adaptive spatial-attention dropout method to facilitate temporal correspondence learning from 2D videos. Notably, we find that our DropMAE achieves better downstream tracking performance than the image-based MAE, while using $50\%$ less pre-training time. Experiments on 6 downstream tracking tasks across 13 benchmarks demonstrate the effectiveness of DropMAE in various tracking applications. We expect our DropMAE to serve as a general pre-trained backbone for various tracking tasks, and inspire more pre-training work in the tracking community. \jimmyy{Future work will explore extending DropMAE to various model architectures (e.g., state space models), tracking paradigms (e.g., autoregressive tracking) and more efficient pre-training strategies.}


\vspace{-0.3cm}
\section{Acknowledgment}
This work was supported by grants from the Research Grants Council of the Hong Kong Special Administrative Region, China (Project No. CityU 11215820) and City University of Hong Kong (Project No. 7030010).


\vspace{-0.3cm}
\small 
\bibliographystyle{IEEEtran}
\bibliography{egbib}

\end{document}